\pdfoutput=1

\documentclass[11pt]{article}

\usepackage[final]{acl}

\usepackage{times}
\usepackage{latexsym}
\usepackage[T1]{fontenc}
\usepackage[utf8]{inputenc}
\usepackage{microtype}
\usepackage{inconsolata}
\usepackage{graphicx}
\usepackage{hyperref}
\usepackage{url}

\usepackage{graphicx}
\usepackage{cleveref}
\usepackage{xcolor}
\usepackage{enumitem} %

\usepackage{listings}
\lstdefinestyle{mystyle}{
    basicstyle=\ttfamily\footnotesize, %
    breaklines=true, %
    breakatwhitespace=false, %
    frame=single, %
    backgroundcolor=\color{lightgray!20}, %
}
\newcommand{\aiannotator}{AI annotator}

\newcommand{\aiannotators}{AI annotators}
\newcommand{\Aiannotators}{AI Annotators}
\newcommand{\llm}{LLM}

\newcommand{\resultfigwidth}{0.43\textwidth}

\newcommand{\change}[1]{\texorpdfstring{{#1}}{{#1}}}
\newenvironment{longchange}{}{\ignorespacesafterend}

\title{Can External Validation Tools Improve\\ Annotation Quality for LLM-as-a-Judge?}

\author{
Arduin Findeis\textsuperscript{$1$$*$$\dagger$}, 
Floris Weers\textsuperscript{$2$}, 
Guoli Yin\textsuperscript{$2$}, 
Ke Ye\textsuperscript{$2$}, 
Ruoming Pang\textsuperscript{$2$}, 
Tom Gunter\textsuperscript{$2$$\ddagger$} \\
\textsuperscript{$1$}University of Cambridge,
\textsuperscript{$2$}Apple
}

\begin{document}

\maketitle
\renewcommand{\thefootnote}{\fnsymbol{footnote}}
\footnotetext[1]{Work done during internship at Apple.}
\footnotetext[2]{\texttt{arduin.findeis@cst.cam.ac.uk}}
\footnotetext[3]{\texttt{tom\_gunter@apple.com}}
\renewcommand{\thefootnote}{\arabic{footnote}}
\newcommand{\codeurl}{\href{https://github.com/apple/ml-agent-evaluator}{github.com/apple/ml-agent-evaluator}}

\begin{abstract}
Pairwise preferences over model responses are widely collected to evaluate and provide feedback to \emph{large language models} (LLMs). Given two alternative model responses to the same input, a human or AI annotator selects the \emph{``better''} response. This approach can provide feedback for domains where other hard-coded metrics are difficult to obtain (e.g., chat response quality), thereby helping model evaluation or training. %
However, for some domains high-quality pairwise comparisons can be tricky to obtain - from AI \emph{and} humans. For example, for responses with many factual statements, annotators may disproportionately weigh \emph{writing quality} rather than underlying facts.
In this work, we explore augmenting standard AI annotator systems with additional tools to improve performance on three challenging response domains: \emph{long-form factual}, \emph{math} and \emph{code} tasks. We propose a \emph{tool-using agentic system} to provide higher quality feedback on these domains. 
Our system uses web-search and code execution to ground itself based on \emph{external validation}, independent of the LLM's internal knowledge and biases. We provide extensive experimental results evaluating our method across the three targeted response domains as well as general annotation tasks, using \emph{RewardBench} (incl. \emph{AlpacaEval} and \emph{LLMBar}), \emph{RewardMath}, as well as three new datasets for domains with saturated pre-existing datasets.
Our results indicate that external tools can indeed improve \aiannotator{} performance in many, but not all, cases. More generally, our experiments highlight the sensitivity of  \aiannotator{} performance to simple parameters (e.g., prompt) and the need for improved (non-saturated) annotator benchmarks. 
We share our code at \codeurl{}.
\end{abstract}

\begin{figure}[ht!]
\begin{center}
\includegraphics[width=0.47\textwidth]{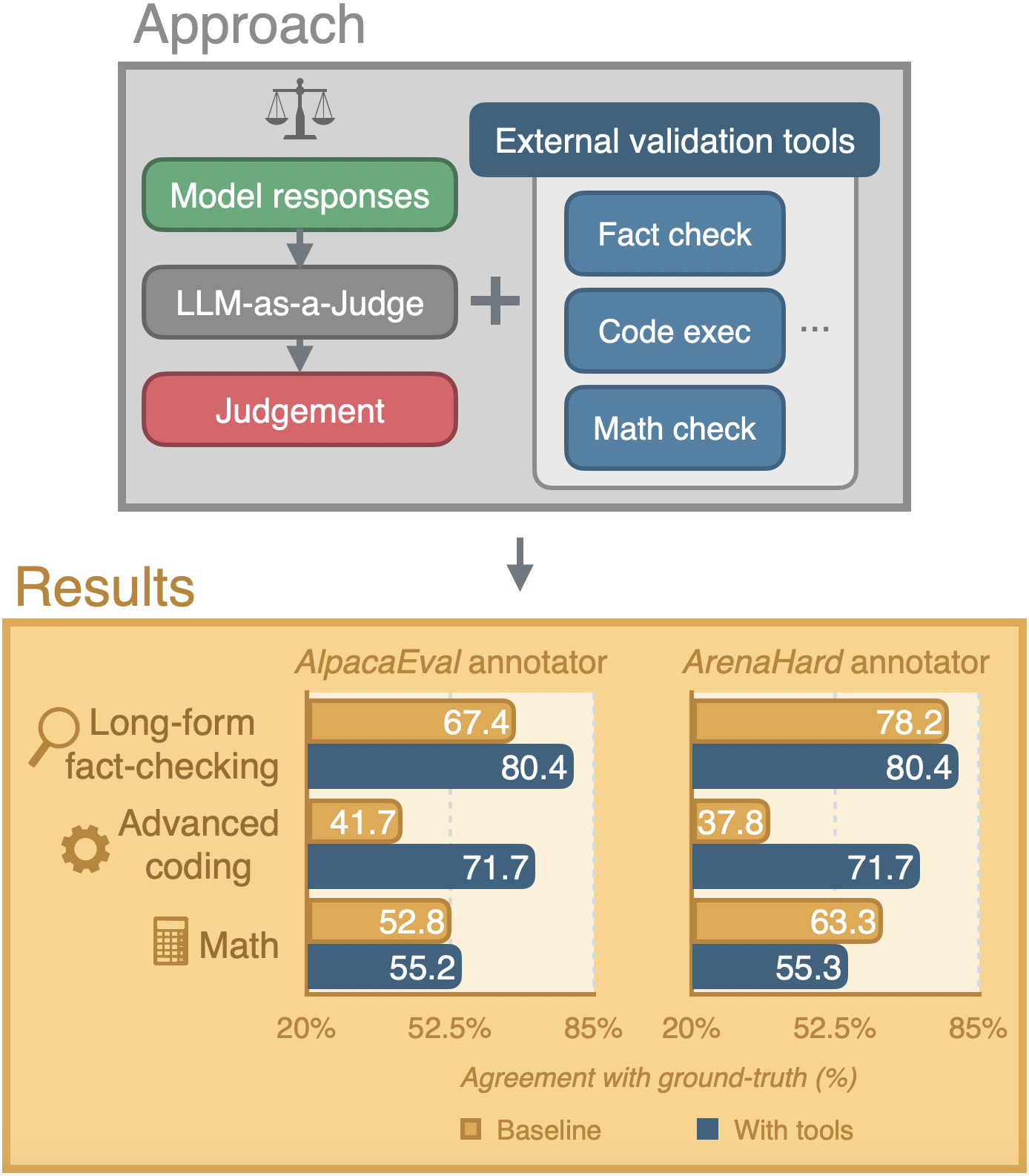}
\end{center}
\caption{\textbf{Summary of our approach and results: We extend standard LLM-as-a-Judge baselines with external validation tools based on web-search and code execution.} We observe that the resulting system is often, but not always, able to improve performance (measured as agreement with ground-truth annotations) across a range of response domains that are typically challenging for LLM-as-a-Judge systems: (1) \emph{long-form factual}, (2) \emph{advanced coding}, and (3) \emph{math} responses. Results with popular AlpacaEval (2.0) and ArenaHard annotators shown, see \Cref{sec:results} for full results.}
\label{fig:abstract_summary}
\end{figure}

\begin{figure*}[ht]
\begin{center}
\includegraphics[width=1\textwidth]{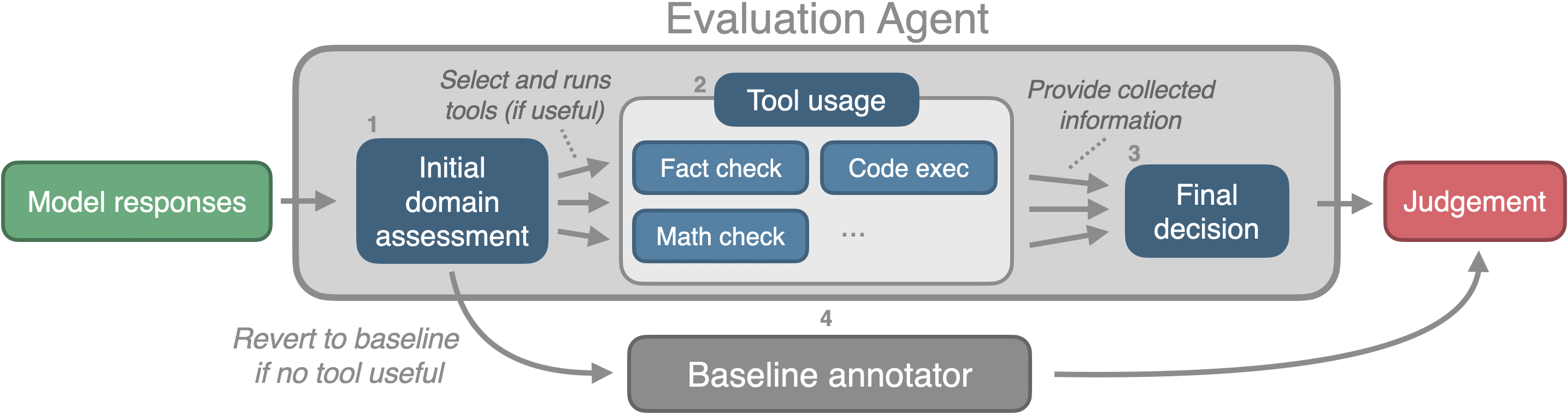}
\end{center}
\caption{\textbf{Overview of our tool-using \aiannotator{} architecture, referred to as \emph{Evaluation Agent}.} In the (1)~\emph{initial domain assessment} the appropriate tools are selected for each response (e.g., for a wiki-style text the fact check tool); then, in (2) \emph{tool usage}, each selected tool is run and the tool outputs are combined into a single prompt to make a (3) \emph{final~decision}. If none of the tools are selected (i.e., no tool deemed useful), the agent instead reverts and returns an annotation from the (4) \emph{baseline annotator} (e.g., AlpacaEval~2.0).}
\label{fig:agent_overview}
\end{figure*}

\section{Introduction}

Pairwise feedback is widely used to understand \llm{} performance on complex tasks that more traditional benchmarks fail to measure well. Given a prompt and two possible responses, the annotator decides which response is \emph{"better"}. This pairwise judgement can be used for \emph{evaluation} (e.g., Chatbot Arena \citep{chiang2024ChatbotArenaOpen}) or to provide feedback for \emph{training} (e.g., via RLHF \citep{stiennon2020LearningSummarizeHumana,ouyang2022TrainingLanguageModels} or DPO \citep{rafailov2023DirectPreferenceOptimization}). Either human or AI annotators, also referred to as \emph{LLM-as-a-Judge}, are used to collect such feedback. Human annotations are often considered higher quality but more expensive.

\emph{Both human and AI annotations have notable limitations:} AI annotators have been observed to be susceptible to a number of biases, including changing preference based on superficial features like \emph{response order} \cite{zheng2023JudgingLLMasaJudgeMTBench} or \emph{response length} \cite{dubois2024LengthControlledAlpacaEvalSimple}. Whilst possibly providing higher quality annotations than AI annotators, human annotators also have known limitations. For example, human annotators have been observed to let their assessment of truthfulness be affected by responses' assertiveness \citep{hosking2024HumanFeedbackNot}.

In certain domains, obtaining high-quality annotations is \emph{particularly challenging}: for responses containing \emph{long-form factual}, \emph{advanced coding} and \emph{math} content both AI and (many) human annotators struggle to provide reliable annotations \citep{zheng2023JudgingLLMasaJudgeMTBench}. Annotating responses in these domains requires expertise and careful deliberation, challenging to achieve for human annotators in a limited amount of time. AI annotators may be less "time-constrained" but nevertheless due to known reliability issues (e.g, hallucinations, limited basic arithmetic) often fail to provide high quality annotations in these domains \citep{yang2023gptsolvemathematicalproblems}.

In this work, we aim to explore improving the annotation quality of widely used AI annotators on these challenging domains by augmenting the annotators with tools that can \emph{externally validate answers}. We enable responses to be fact-checked using \emph{web-search}, or verified using \emph{code execution}. Our setup is illustrated in \Cref{fig:abstract_summary,fig:agent_overview}. In particular, we make the following contributions:

\begin{enumerate}[wide, labelwidth=0pt, labelindent=0pt, left=3pt,itemsep=-0.1em]
    \item \textbf{Extensible framework for using tools with existing \aiannotators{}}. We introduce a new framework that enables the integration of new tools on top of existing \aiannotators{} to improve annotation quality in certain domains using external validation. Our framework \change{is \emph{agentic} in the sense that an LLM} assesses the response domain and plans the optimal tool usage accordingly.\footnote{See \Cref{app:agenttermdiscussion} for further discussion of our use of the term \emph{agentic}.} We provide a number of initial tool implementations: (1) a \emph{long-form fact checking} tool based on the \emph{Search Augmented Fact Evaluation} (SAFE) method by \citet{wei2024LongformFactualityLarge}; (2) a \emph{code check} tool built on OpenAI's code interpreter API; and (3) a \emph{math check} tool similarly built on code execution. We open-source the framework's code.\footnote{\codeurl{}}
    \item \textbf{Comprehensive experimental results evaluating our framework's capabilities.} We evaluate our framework's effectiveness across a wide range of tasks including newly created datasets as well as well-established benchmarks. We compare our method to a number of popular state-of-the-art \aiannotators{}, including the annotators underlying \emph{AlpacaEval 2.0}~\citep{dubois2023AlpacaFarmSimulationFramework}, and \emph{ArenaHard}~\citep{li2024CrowdsourcedDataHighQuality}.%
\end{enumerate}

\section{Problem: Pairwise Feedback on Complex Tasks}
\label{sec:problem}

For many task domains, pairwise feedback can be easier to obtain than absolute metrics. Nevertheless, for some domains even a relative pairwise judgement can be difficult to collect --- from both human \emph{and} AI annotators. In this work, we consider three particularly challenging response domains: tasks that require model responses with (1) \emph{long-form factual}, (2) \emph{advanced coding} or (3) \emph{math} content. For such tasks, even a relative judgement requires robust understanding of the task domain, and, for human annotators, careful deliberation. For example, judging code without understanding the relevant syntax may force an annotator (AI or human) to revert to higher level features such as style – that may not fully correlate with ground-truth preferences. Similarly, when comparing responses with a large number of factual statements, an annotator may easily miss a single incorrect factual statement --- instead possibly again relying on writing style to make a judgement. \change{At the same time, annotators only judging according to factual or functional correctness may miss other response traits (e.g., readability) distinguishing a merely \emph{correct} from an \emph{excellent} response.}

In the pairwise setting, annotators are typically evaluated based on their \emph{agreement}\footnote{Note that other works (e.g, \citep{bavaresco2024LLMsInsteadHuman}) use Cohen's kappa. However, to retain consistency and comparability with our primary benchmark RewardBench \citep{lambert2024RewardBenchEvaluatingReward}, and for better interpretability, we report all our results using the more common accuracy (agreement) metric. With the agreement metric, random performance is $\sim$50\%.} with ground-truth annotations on datasets, where such annotations are either available by construction or created by human annotators \citep{lambert2024RewardBenchEvaluatingReward}. This agreement is equivalent to the accuracy of the binary classification task of predicting the correct ranking for each response pair. In this setting, the goal of pairwise feedback annotation is to \emph{maximise} the agreement with ground-truth annotations.
In general, for many response pairs there is ambiguity regarding which response is better --- especially for domains with known disagreements such as political preferences \citep{kirk2024prism}. To improve the reliability of our evaluation, we primarily test on response pairs where experts agree on the preference and avoid more contentious topics.

\section{Method: \Aiannotators{} with Tools for External Validation}
\label{sec:method}

\begin{figure*}[ht!]
\begin{center}
\includegraphics[width=1.\textwidth]{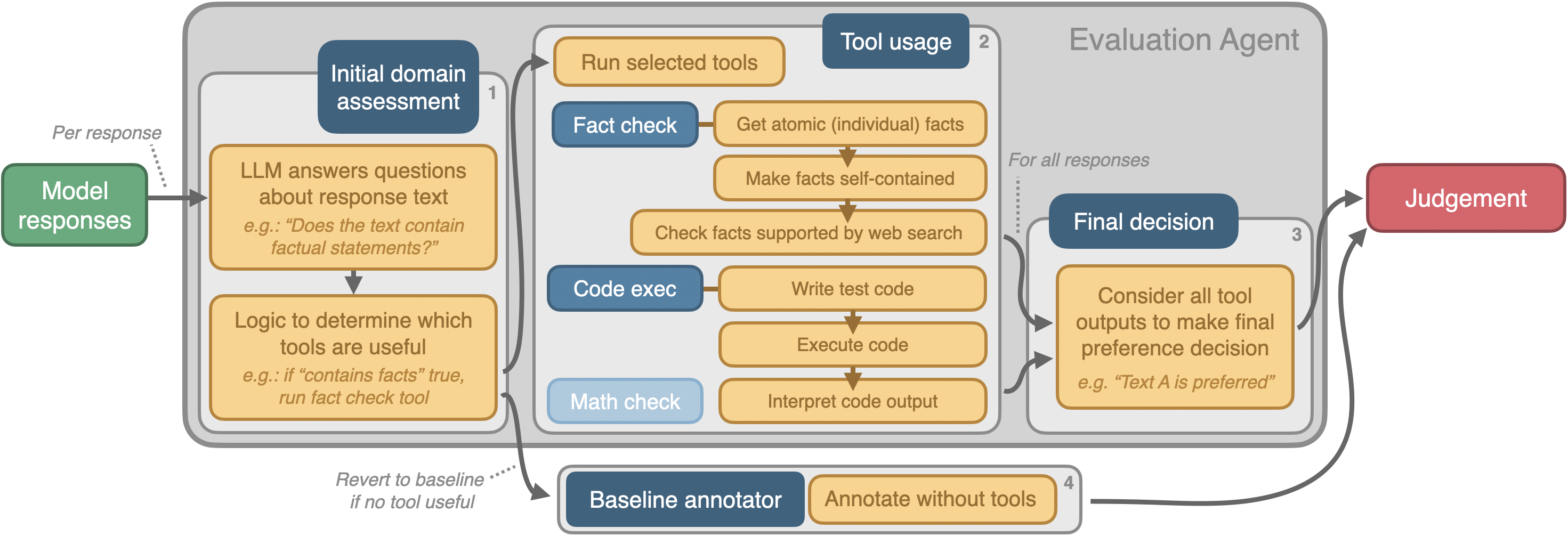}
\end{center}
\caption{\textbf{Detailed overview of our evaluation agent:} the model responses are first processed by the (1) \emph{initial domain assessment}, where an LLM is prompted to answer questions about the response text. In (2) \emph{tool usage}, each tool that is deemed useful in Step (1) is run. Initially, available tools include \emph{fact check}, \emph{code exec} and \emph{math exec}. The first tool is based on web-search, the latter two tools on a code interpreter. Finally, in the (3) \emph{final decision} step, an LLM makes a final preference decision considering all tool outputs across responses together. If the (1) \emph{initial domain assessment} finds no useful tool, the entire approach reverts back to the (4) \emph{baseline annotator}'s judgement.}
\label{fig:tools_overview}
\end{figure*}

We introduce a new framework for augmenting existing \aiannotators{} with tools – grounding their annotations in the real world with external validation. The general functioning of our framework is illustrated in \Cref{fig:agent_overview}. Our goal is to improve the performance of \aiannotators{} on a specific set of \emph{target domains}: responses containing \emph{long-form factual}, \emph{advanced coding} and \emph{math} content. To achieve this annotation quality improvement, we leverage external validation via tools built on \emph{web search} and \emph{code execution}. At the same time, we want to avoid reducing performance on other \emph{non-target} domains. We use an agentic setup to determine when each tool gets used, letting an underlying \llm{} assess the domain of the response considered and thereby which tool would be most useful. To avoid regression on non-target domains, our agentic framework reverts back to a baseline annotator whenever the responses are assessed to be outside the domain of all available tools. We build on \emph{structured output} throughout our pipeline to reduce the parsing error rate. Instead of plain text responses, structured output forces the model to return JSON-formatted outputs. With this approach, each LLM call is not only configured by a single prompt message but also by the JSON format and description of the requested output.

Shown in \Cref{fig:tools_overview}, our approach consists of three distinct parts: (1) \emph{initial domain assessment}, determining which tools to use (if any); (2) \emph{tools}, running the selected tools for each response; and (3) \emph{final decision}, creating a final preference judgement based on all outputs. If the first step (\emph{initial domain assessment}) determines that no tools would be helpful, our approach alternatively skips steps (2) and (3). Instead, we revert to the (4) \emph{baseline annotator}. In the following subsections, we describe each step in more detail. For full transparency, we share the prompts in \Cref{app:prompts} and make the corresponding code publicly available.\footnote{\codeurl{}}

\begin{description}[leftmargin=0cm]
\item[Step 1: Initial domain assessment.] 
The \emph{initial domain assessment} ensures that each tool is only run if the model responses are within a domain where the tool is known to be likely helpful. For example, for the \emph{code execution} tool, the domain assessment ensures that \emph{there is code present in the response}. This assessment helps avoid running tools in scenarios where they are unlikely to help. For each tool, we created a number of questions about a response (e.g., "Whether text might benefit from running code."). For each response, an LLM is prompted with these questions. The LLM's parsed answers determine whether a tool is deemed useful and run – or not. If not a single tool is deemed useful, the agent reverts back to a baseline evaluator. With this setup, our method aims to reduce unnecessary inference costs and to avoid regressing on domains where the tools are not useful. In the tie case, when tools are only considered useful for one of the two texts, our framework reverts back to the baseline with 50\% probability and uses the agent otherwise. \change{Further, clearly separated tool domains in our setup allow integrating a large number of domain-specific tools whilst avoiding adverse effects out-of-domain.}

\item[Step 2: Tool usage.]

If the initial assessment deems one or more tools useful, the respective tools are run. We initially implemented three different tools as part of our extensible framework\change{, chosen to specifically tackle the limitations of LLM-as-a-Judge systems discussed in \Cref{sec:problem}:}

\textbf{Tool A: Fact-checking.} We build on the \emph{Search Augmented Factuality Evaluator} (SAFE) by \citet{wei2024LongformFactualityLarge} to create a fact-checking tool for the pairwise setting. Our fact-checking tool follows similar steps as the original SAFE algorithm: (1) \emph{separating atomic facts}, (2) \emph{making atomic facts self-contained}, and (3) \emph{checking whether self-contained facts are supported by web-search}. Our tool omits the \emph{relevance check} in the original SAFE algorithm. In a pairwise preference setting we consider the truthfulness of all facts relevant, even if the facts are not directly related to the task. The final assessment ultimately decides which factual statements are most relevant. Note that our approach currently relies on the LLM itself to judge the web search results. The method does not currently explicitly verify the information from the web beyond the LLM’s judgement.

\textbf{Tool B: Code execution.} Taken into account existing works that show that compiler/runtime output is a useful signal, we build on top of OpenAI's code interpreter API to create a code-execution tool. For both proposed answers to a prompt, the code-execution tool will verify its correctness using execution feedback. Internally, OpenAI's code interpreter API can create additional unit tests, run multiple execution steps and draw a conclusion. Only the last conclusion is used in the agent's final assessment to determine which response is better.

\textbf{Tool C: Math checker.} Noting that autoregressive language models are not reliable arithmetic engines \citep{yang2023gptsolvemathematicalproblems}, we prompt-constrain our code-execution tool to perform math (and in particular arithmetic) validation on each of the model outputs. As in the case of general code execution, multiple checks may be executed per model output, and the final assessment uses the outcome of these checks to inform its overall decision. \change{We created a separate math checker after preliminary tests indicated a standard code interpreter tool does not transfer well to math annotation settings.}

\item[Step 3: Final assessment.]

In the \emph{final assessment} step, we combine the results of all tools per response alongside the original prompt and response, to ask an \llm{} to make a preference judgement based on all collected information. Critically, this step allows the \llm{} to access the external validation results when making a decision. The \llm{} response to this step provides the final preference judgement (e.g., \emph{"Text A is preferred."}) as well as a chain-of-thought (CoT) reasoning for the judgement (e.g., \emph{"Text A is preferred because [...]"}).

\end{description}

\section{Experimental Results}
\label{sec:results}

\subsection{Datasets}
\label{sec:datasets}

\textbf{Existing datasets.} A number of benchmarks aim to evaluate \aiannotator{} capabilities, notable examples include (subsets of) \emph{AlpacaEval}
~\citep{dubois2023AlpacaFarmSimulationFramework}, \emph{MT-Bench}~\citep{zheng2023JudgingLLMasaJudgeMTBench}, \emph{LLMBar}~\citep{zeng2024EvaluatingLargeLanguage} and \emph{RewardBench} \citep{lambert2024RewardBenchEvaluatingReward}. We use the latter, RewardBench, for our evaluation, as it represents a superset including the other tasks. This benchmark provides a broad coverage of response domains, including \emph{mathematical reasoning}, \emph{code generation} and \emph{general chatbot conversation}. We find that some subsets of the benchmark are fairly saturated: state-of-the-art LLM-as-a-judge systems already achieve close to 100\% agreement with the ground-truth annotations (see \Cref{app:rewardbenchdiscussion} for discussion). Thus, to effectively evaluate improvements in these domains, we created new pairwise datasets.

\textbf{New pairwise datasets.} We extend RewardBench by adapting existing, more challenging (previously non-pairwise) datasets to the pairwise setting. \Cref{app:dataset_examples} contains examples from each dataset introduced below.

\begin{enumerate}[wide,,itemsep=-0.1em, labelwidth=!, labelindent=0pt, leftmargin=5pt]
\item \textbf{Long-form fact checking: LongFact pairwise.} We create a dataset of response pairs, where responses vary in long-form factual correctness, using the LongFact prompt dataset by \citet{wei2024LongformFactualityLarge}. We use OpenAI's \emph{gpt-4o-mini-2024-07-18} model to generate two long-form factual responses for each prompt. \change{We then manually introduce factual errors into one of the responses.} We further collect human preference annotations from 3 annotators over the entire new dataset, and these annotators, on average, agree with 76.83\% of those ground-truth annotations when \emph{not} selecting a tie. 18\% of the average human annotations are ties. \change{Full details on the data generation process are available in \Cref{app:data_gen}}.

\item  \textbf{Challenging coding: APPS competition pairwise.} From the original APPS dataset \citep{hendrycks2021MeasuringCodingChallenge}, we create a pairwise response dataset to evaluate the ability to determine code correctness. The APPS benchmark contains coding problems, unit tests and Python ground-truth solutions for most problems. We take the ``competition'' subset, arguing it is these harder problem/solution combinations that are tricky to evaluate correctly. We only keep samples that contain a ground-truth solution, leaving us with 310 items. We then use GPT-4-0613 to generate solutions to the problems, \change{until} we have failing solutions for all 310 items.

\item \textbf{Challenging maths: GSM8k hard pairwise.} We select a ``hard'' subset of the GSM8k \citep{gsm8k} dataset by keeping the 11\change{6}  examples that GPT-4o is unable to solve. For each example, we generate pairwise responses by keeping both the ground-truth answer and the incorrect answer that GPT-4o provided. We also conducted a detailed analysis of validity of the GSM8k datapoints used, shared in \Cref{app:gms8k_analysis}.

\end{enumerate}

We additionally create a pairwise response dataset where responses vary in \emph{short-form} factual correctness using the TruthfulQA dataset%
by \citet{lin2022TruthfulQAMeasuringHow}. Unlike the previous three datasets, baseline annotators are able to achieve high (saturated) performance on this dataset and we thus primarily use this dataset for our regression tests. \change{Further,} unlike the long-form responses in our LongFact pairwise dataset, responses in this dataset are typically between a single word and single sentence long, relating to a single fact. \change{See \Cref{app:data_gen} for full data generation details.}

\subsection{Baseline Annotators}

We compare our method to two popular \aiannotator{} configurations that are widely used in academic and industry settings, and may be considered \emph{state-of-the-art}: (1) the widely-used \emph{AlpacaEval 2.0}\footnote{Config. name: \texttt{weighted\_alpaca\_eval\_gpt4\_turbo}.} annotator by \citet{dubois2023AlpacaFarmSimulationFramework} using \emph{GPT-4-Turbo}, logprob parsing to extract annotations; and (2) the \emph{ArenaHard} annotator by \citet{li2024CrowdsourcedDataHighQuality} using more extensive annotation instructions (including asking the model to craft its own response) and string parsing; 
We further share results using two minimalist \aiannotators{} that simply ask the underlying \llm{} to \emph{"select the better"} text, powered by GPT-3.5-Turbo and GPT-4o. Perhaps surprisingly, we find that the simple annotator powered by GPT-4o performs competitively on many datasets considered in our experiments.
We report all results based on 5 seeds (unless otherwise specified), showing the mean with standard deviation as error bars. When reporting the agent results across different baselines, we use the same 5~seeds of the agent Steps 1-3, only changing the underlying baseline results (Step 4). This setup notably reduces the cost of our experiments as agent steps required the majority of inference compute.

\subsection{Results on Target Domains}

\change{In this section we show results on the targeted domains: long-form factual, code and math tasks.}

\subsubsection{Long-Form Fact-Checking}
\label{sec:exp:factcheck}

\begin{figure}[t]
\begin{center}
\includegraphics[width=\resultfigwidth]{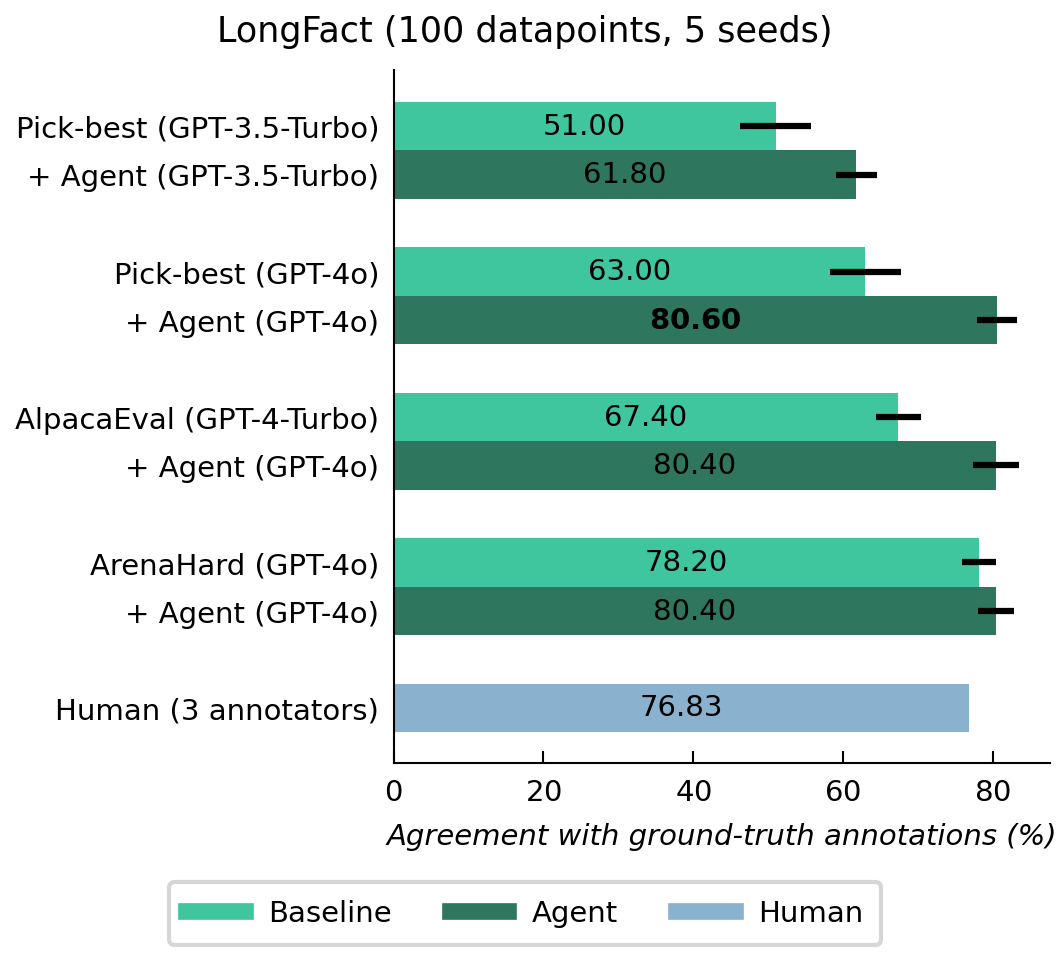}
\end{center}
\caption{\textbf{Long-form fact checking results on LongFact pairwise data.} We augment multiple baseline annotators (\emph{light green}) with our evaluation agent framework (\emph{dark green}) and observe that our agents have higher average agreement with ground-truth annotations across baselines. The effect is most pronounced for simpler baselines, including when agent and baseline are based on the less capable GPT-3.5-Turbo model. We also collect non-expert human annotations (\emph{blue}) for the same dataset, and observe that, when making a non-tie judgement, human annotators have higher disagreement with the ground-truth than our best agent evaluators.}
\label{fig:factchecking_main}
\end{figure}

We evaluate our method on data pairs that require long-form fact checking using the \emph{LongFact pairwise} dataset introduced in \Cref{sec:datasets}. \Cref{fig:factchecking_main} illustrates our results on this dataset.

\textbf{Observation 1: Our external validation tools can help \aiannotators{} improve performance annotating long-form factual responses.} In \Cref{fig:factchecking_main} we observe that, across all evaluated baselines, augmenting any baseline with our fact-checking agent helps improve the overall agreement with the ground-truth annotations on this dataset. Whilst the contrast is most pronounced with simpler baselines (e.g., for GPT-4o \emph{pick-best baseline}, 63\% vs 81\%), the effect is present across all baselines, including ArenaHard (78\% vs 80\%). 

\textbf{Observation 2: For baseline annotators, configurations such as prompt have a strong impact on the downstream performance on long-form fact checking (jumping from 63\% to 78\% for GPT-4o).} We observe a jump in agreement between the \emph{pick-best} and \emph{ArenaHard} baseline annotators, both powered by GPT-4o. The only difference between these annotators is the prompt and answer parsing used. The \emph{pick-best} annotator uses a simple prompt asking for the better answer, either text A or B. The \emph{ArenaHard} annotator uses an extensive prompt, including asking the \llm{} to create its own response for comparison. This observation indicates that for this type of factual task the exact choice of \aiannotator{} configuration is critical, with the \emph{ArenaHard} configuration performing the best amongst the tested baselines. 

\textbf{Observation 3: Our agents' agreement with our ground-truth annotations is higher than human annotators' on long-form factual responses.} This effect holds for all agents based on baselines with GPT-4-style models. \citet{wei2024LongformFactualityLarge} similarly report their method sometimes outperforming non-expert human annotators. Intuitively, it seems plausible that human annotators may be affected by time limits and fatigue – unlike our agent. \cite{hosking2024HumanFeedbackNot} similarly observe that human annotators' perceived rate of factual errors can be skewed by the assertiveness of a model response, indicating that human annotators may not always consider factual errors sufficiently.

\subsubsection{Math-Checking}
\label{sec:exp:mathcheck}

\begin{figure}[t]
\begin{center}
\includegraphics[width=\resultfigwidth]{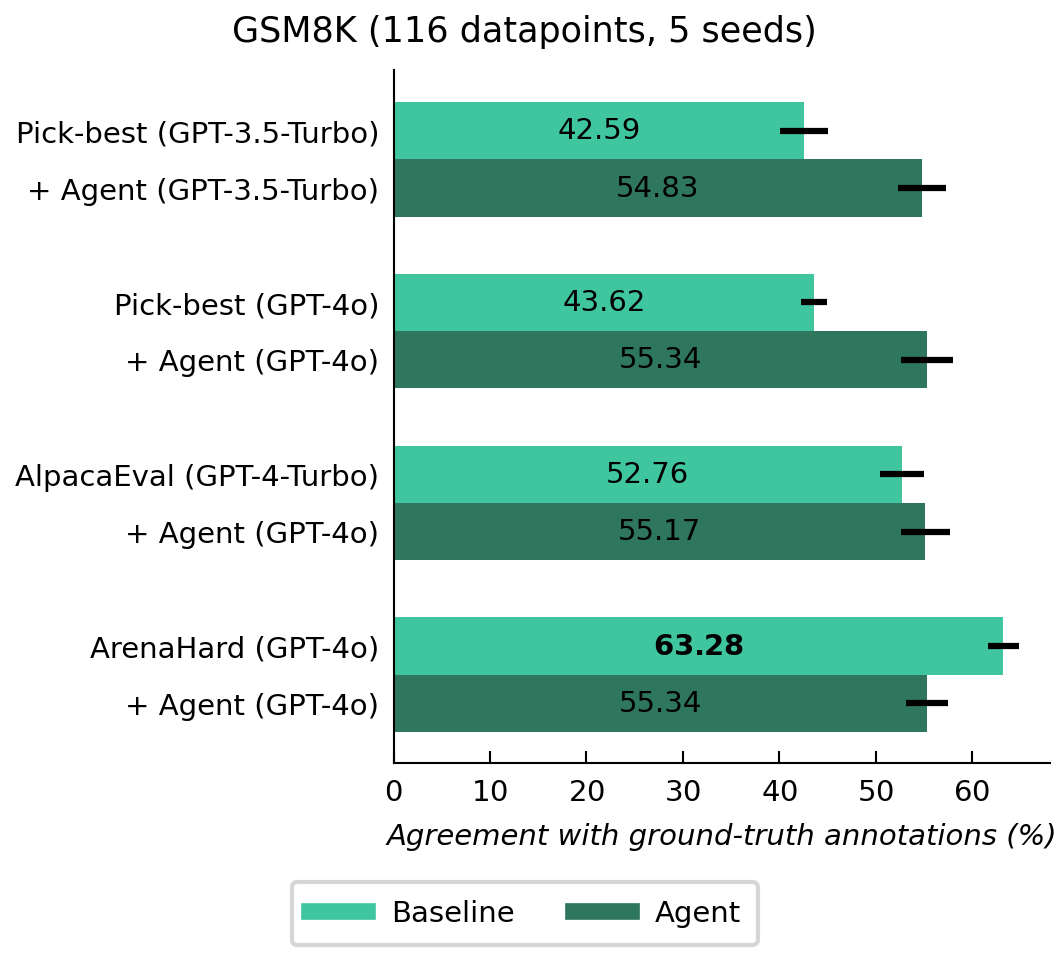}
\end{center}
\caption{\textbf{Results annotating responses on our pairwise set of mathematical tasks based on GSM8k}. We observe that our method improves performance over some baselines,  but the overall level of agreement remains relatively low (around 56\%). Further work is needed to improve the models capability to leverage code execution fully in a math context. %
}
\label{fig:math}
\end{figure}

We further evaluate our method on annotating solutions to advanced mathematics tasks, via the \emph{GSM8k hard pairwise} dataset introduced in \Cref{sec:datasets}, the results are shown in \Cref{fig:math}. 

\textbf{Observation 4: Our agents are able to outperform some, but not all, baselines on hard math annotation tasks based on GSM8k.} We observe that only some augmented baseline annotators are able to improve their performance. In particular, the \emph{ArenaHard} annotator is notably able to outperform all agent-based methods on this task.
\begin{longchange}
This result indicates that more complex prompting methods (in terms of token usage and code), such as our framework, do not necessarily always improve annotator performance over (relatively) less complex methods, such as ArenaHard. 
\end{longchange}
Future work may be able further improve our method's ability to use code execution in a math context. 

To further evaluate our method's ability to improve math performance, we additionally conduct an experiment on the RewardMATH dataset by \citet{kim2024EvaluatingRobustnessReward}. Unlike on the GSM8k dataset, we observe our method outperforming the ArenaHard baseline on RewardMATH. The detailed results are shared in \Cref{app:rewardmath}.

\subsubsection{Code-Execution}
\label{sec:exp:codeexec}

\begin{figure}[t]
\begin{center}
\includegraphics[width=\resultfigwidth]{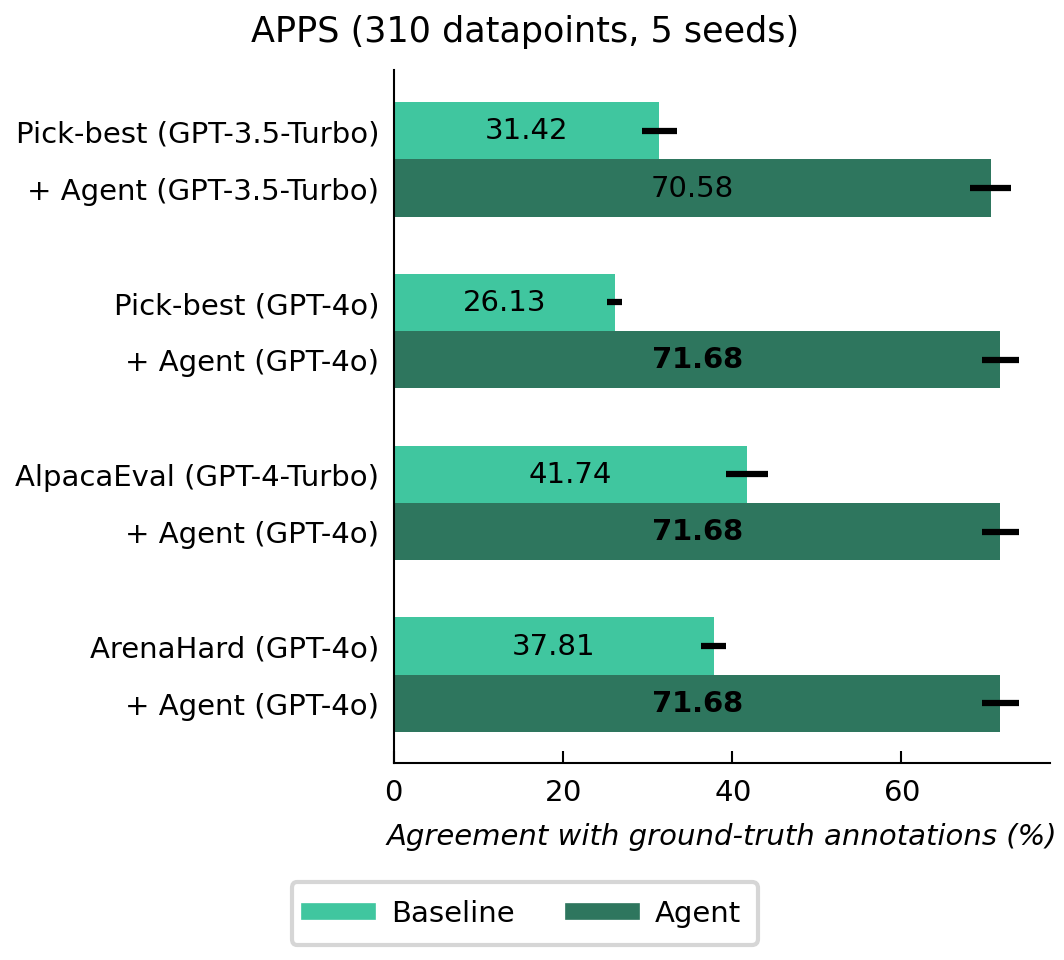}
\end{center}
\caption{\textbf{Results on our pairwise dataset of responses to advanced coding tasks from the APPS  dataset} \citep{hendrycks2021MeasuringCodingChallenge}. We observe a notable improvement of our method over the baseline results, even for the otherwise less capable models GPT-3.5-Turbo.}
\label{fig:coding}
\end{figure}

Finally, we evaluate our method's ability to improve capabilities in annotating advanced coding tasks using our pairwise coding dataset based on the \emph{APPS} dataset by \cite{hendrycks2021MeasuringCodingChallenge}. The results are shown in \Cref{fig:coding}. 

\textbf{Observation 5: Our method is able to notably improve the baseline performance on annotating the APPS advanced coding responses.} Across all baselines, our agent-based approach is able to notably improve annotation performance. This improvement holds both for the less capable GPT-3.5-Turbo model (31\% baseline vs 71\% agent) as well as the \emph{ArenaHard} annotator that performs strongly on other tasks (38\% baseline vs 72\% agent). 

\textbf{Observation 6:  Baseline annotators \change{perform worse than random on APPS dataset}.} Based on the construction, there may be slight style differences between correct (pre-existing ground-truth solutions) and incorrect responses (GPT-4 generated \emph{incorrect} code), see examples in \Cref{app:dataset_examples}. We observe that all baseline annotators have a bias towards the incorrect GPT-4 responses, preferring only 26\% to 42\% of correct responses. This effect may possibly be explained with self-enhancement bias \change{\citep{panickssery2024LLMEvaluatorsRecognize,stureborg2024LargeLanguageModels}}. Our agent method using code execution does not show such misaligned preferences.

\subsection{Results Outside of Target Domains (Out-of-Domain)}
\label{sec:ood_results}

\begin{figure}[t]
\begin{center}
\includegraphics[width=\resultfigwidth]{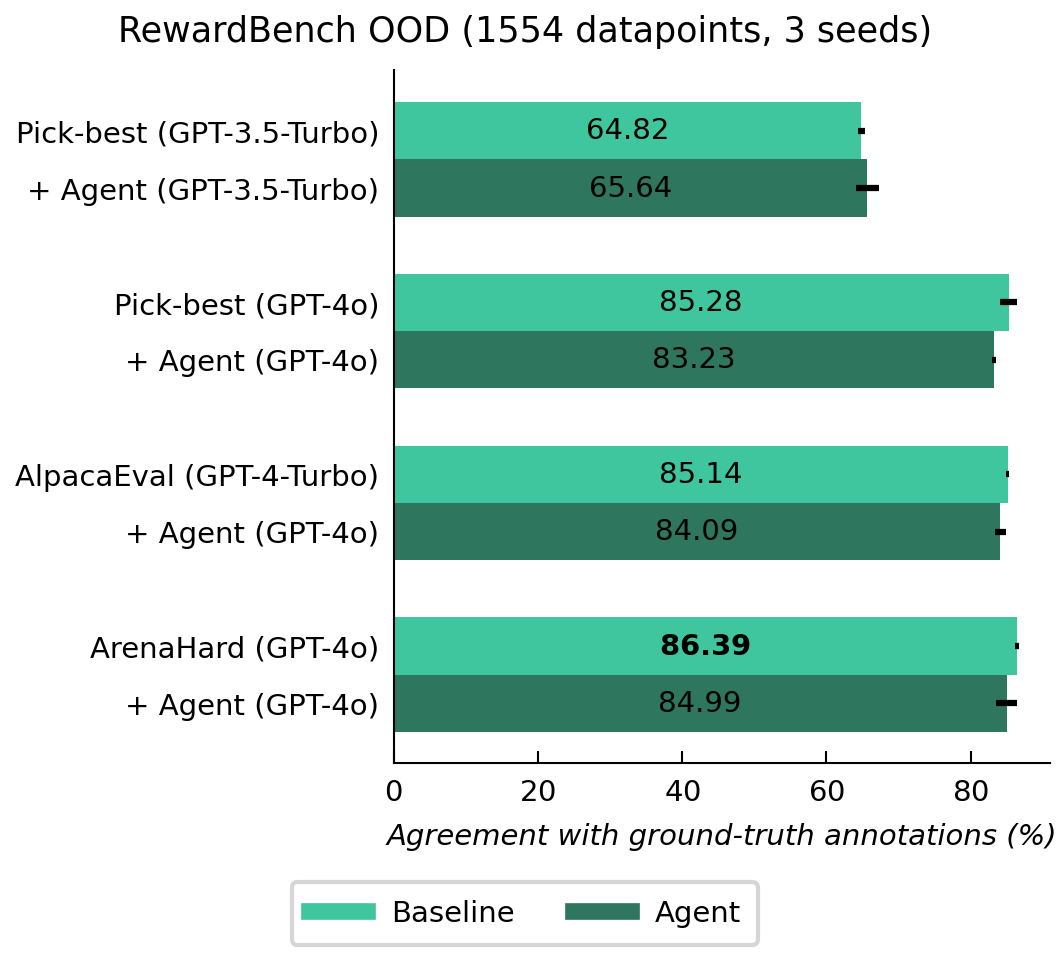}
\end{center}
\caption{\textbf{General out-of-domain annotation capabilities result based on RewardBench} \citep{lambert2024RewardBenchEvaluatingReward}. We observe that our agent achieves similar performance to the baseline annotator across these tasks --- at worst seeing a reduction of $\sim$2\% in agreement.}
\label{fig:outofdomain_results}
\end{figure}

In practice, an \aiannotator{} may encounter response pairs from across a variety of task domains – including domains not intended to be addressed by our method. A good \aiannotator{} should be able to work across domains, as filtering data may not always be feasible or sufficiently effective. Thus, we go beyond the domain-specific capability improvements shown in \Cref{sec:exp:factcheck,sec:exp:codeexec,sec:exp:mathcheck} and also evaluate our method's performance on \change{RewardBench} tasks that are out-of-domain for our tools\footnote{\change{This out-of-domain dataset includes the \emph{Chat}, \emph{Chat Hard} and \emph{Safety} RewardBench categories.}}. \emph{In this general scenario we would not expect performance improvements with our method} but aim for minimal performance regression – as our tools are not built to help (or activate) on most of these tasks. \Cref{fig:outofdomain_results} shows our results on these tasks.

\textbf{Observation 7: On out-of-domain tasks from Rewardbench there are minimal performance reductions using our approach with any tested baseline.} The agreement reductions are less than 2\% for all tested baselines. For the GPT-3.5-Turbo-based agent we even observe a slight improvement. Future work may be able to refine the initial assessment to further reduce this gap.

\textbf{Further analysis of agent performance.} To better understand why performance sometimes reduces slightly in \Cref{fig:outofdomain_results} with the agent, we conduct further analysis of the agents' performance. First, we investigate how often agent reverts to the baseline annotator: for our out-of-domain experiments our agents revert for 73.9\% of datapoints, for the in-domain experiments (LongFact, GSM8k and APPS) our agents only revert for between 0.2\% to 2.2\% of datapoints. Whilst our domain assessment correctly identifies the in-domain tasks, further adaptions to the assessment may help further reduce the activation out-of-domain. 
We further manually inspect the failure reasons for 30 datapoints where the agent fails to annotate correctly. We observe that for 9 out of 30 examples the agent \emph{chooses the wrong tool} for the task. For example, the agent sometimes uses the fact-checking tool when a refusal response should be selected for safety-reasons. Further, for 18 out of the 30 datapoints, we observe that tool-use does not fix existing capability issues: both with and without tools the annotator makes the wrong decision. For 6 of these 18 datapoints, the previously described safety scenario also applies. Additional details of the manual inspection are provided in \Cref{app:ood_tool_activation,app:ood_failure_manual_inspection}.

\textbf{Results on adjacent domains.} Further, we specifically evaluate our results on domains closely adjacent to our main focus domains: short-form fact checking (TruthfulQA pairwise), simple coding tasks (RewardBench HumanEval pairwise) and general math problems (RewardBench PRM pairwise). These domains are already quite well solved by state-of-the-art \aiannotators{}. Thus, as with the general out-of-domain results, we would not expect any notable improvements but aim to demonstrate \emph{limited performance regressions}. We observe two opposing effects: for the short-form fact checking and simple maths our approach is consistently able to improve performance, whereas for simple coding tasks the annotation performance decreases (reduction of up to 9\%, see \Cref{app:fig:rewardbench_code_results}). One possible explanation may be that the very high baseline performance on HumanEval (above 97\% for GPT-4-style models) may be reduced by additional noise due to code execution pipeline. \Cref{app:adjacant_domain_results} includes detailed results for these adjacent domain experiments.

\section{Related Work}
\label{sec:related_work}

\textbf{Pairwise \aiannotators{}.} As human annotations are costly and time-intensive, extensive work has been done to explore the use of \emph{\aiannotators{}} as an alternative. Works such as \emph{LLM-as-a-judge} \citep{zheng2023JudgingLLMasaJudgeMTBench}, \emph{AlpacaEval} \citep{dubois2023AlpacaFarmSimulationFramework} and \emph{G-Eval} \citep{liu2023GEvalNLGEvaluation} popularized AI annotators in the context of evaluation. The \emph{ArenaHard} annotator is another popular choice \citep{li2024CrowdsourcedDataHighQuality}. Various efforts have also explored the use of AI annotators for generating training data, such as \emph{constitutional AI} \citep{bai2022ConstitutionalAIHarmlessness}. This line of work is also known as \emph{reinforcement learning from AI feedback} (RLAIF) \citep{lee2024RLAIFVsRLHF}.

\textbf{\aiannotator{} problems.} A number of biases have been observed in \aiannotators{}, for example (1) \emph{length bias} \citep{zheng2023JudgingLLMasaJudgeMTBench,dubois2024LengthControlledAlpacaEvalSimple}, where annotators prefer more verbose outputs (even when not corresponding to human preference); (2) \emph{position bias} \citep{zheng2023JudgingLLMasaJudgeMTBench}, where the model's annotation affected by order in which they are shared with the model; and (3) \emph{self-enhancement bias} \citep{panickssery2024LLMEvaluatorsRecognize, stureborg2024LargeLanguageModels}, where preferred responses have high probability under judge model's distribution.

\textbf{Augmented AI evaluators.} Given the known limitations of basic \aiannotators{}, various \emph{augmentations} of such annotators have been explored. \change{\citet{li2024ToolAugmentedRewardModeling} explore the use of external validation tools to improve the performance of a reward model (RM), in a framework named \emph{Themis}. Similar to our work, the tools considered include code interpreter and web search tools. However, Themis requires a language model with customized architecture and fine-tuning---preventing the use of Themis with the latest state-of-the-art closed-source models.} We conducted experiments applying Themis to the datasets considered in our work with limited success, the results are discussed in \Cref{app:themis}. \citet{dubois2024LengthControlledAlpacaEvalSimple} propose augmenting AI annotators to be length-controlled using a generalized linear model to address the widely observed length bias. Others explore using multiple \aiannotators{} simultaneously to improve performance \citep{verga2024ReplacingJudgesJuriesa,chan2023ChatEvalBetterLLMbasedd}.

In a non-pairwise setting, the \emph{Search Augmented Factuality Evaluator} (SAFE) by \citet{wei2024LongformFactualityLarge}, and prior work FActScore \citep{min2023FActScoreFinegrainedAtomic}, RARR \citep{gao2023RARRResearchingRevising}, Factcheck-Bench \citep{wang2024FactcheckBenchFineGrainedEvaluation}, all aimed to improve the capability of verifying facts within text, such as model responses. \change{\citet{gou2023CRITICLargeLanguage} explore the use of external validation tools to improve \emph{generative} performance, demonstrating improvements for question answering, programming and toxicity reduction tasks.}

\section{Conclusion}

In this work we have presented a novel framework for augmenting \aiannotators{} with tools to externally validate outputs and address existing limitations with AI and human annotations. We compare our method to state-of-the-art and widely used \aiannotators{}, including the \emph{AlpacaEval 2.0} \citep{dubois2023AlpacaFarmSimulationFramework} 
and \emph{ArenaHard} annotator \citep{li2024CrowdsourcedDataHighQuality}. To challenge our method on annotation tasks where the existing datasets appear saturated (coding, math) or little pairwise data exists (long-form factual responses), we created new pairwise datasets, building on \emph{LongFact} \citep{wei2024LongformFactualityLarge}, \emph{GSM8k} \citep{cobbe2021TrainingVerifiersSolve}, and \emph{APPS} \citep{hendrycks2021MeasuringCodingChallenge}.
We evaluate our method's effectiveness across a diverse collection of datasets, including the new datasets and RewardBench \citep{lambert2024RewardBenchEvaluatingReward}. We observe that our external validation-based method often improves baseline annotator performance, \change{with} strongest effectiveness \change{when} annotating \emph{advanced coding} responses but also for \emph{long-form factual} responses, with more mixed results in \emph{advanced math} responses. We conclude that, whilst external validation tools can often improve annotation quality of AI annotator (or \emph{LLM-as-a-Judge}) for certain scenarios, such tools represent a trade-off in terms of complexity and cost. Careful evaluation is required to effectively apply such tools and they may not be the right fit for every use-case.%

 \section{Limitations}

 As discussed in \Cref{sec:ood_results}, our method does (as expected) currently show some regression on some out-of-domain tasks. Thus, in practice, our method's overall usefulness will depend on the domain distribution of the datasets it is applied to. For datasets with a high proportion of datapoints in our target domains, our method is likely able to improve annotation quality. For more out-of-domain datasets any performance improvement will likely be limited.

 Further, as discussed in \Cref{sec:problem}, our experiments are limited to domains with high expert agreement. Domains where expert agreement is not necessarily given are more difficult to target, as the goal of judges is less clearly defined in such a case. This limitation applies to both for our system and LLM-as-a-Judge systems in general.

Potential risks of using our method include over-relying on LLM-as-a-Judge systems rather than human judgements, possibly leading to misaligned models that are overfit to such AI judges. In general, LLM-as-a-Judge methods should be used to complement -- rather than replace -- human judgement.

 More broadly, our results highlight the strong effect that simple configuration parameters, such as prompt and parsing method, can have on annotator performance --- even if the same underlying LLM is used. When considering more technically involved augmentations like our external validation tools, we recommend to also carefully evaluate simpler configurations as an alternative across a wide range of scenarios, as we have done. A robust AI annotator testing pipeline can be critical to determine the right annotator. Concurrent work by \citet{calderon2025AlternativeAnnotatorTest} offers a promising direction for more rigorous statistical tests of annotators.
 We would welcome future work that develops further datasets and methods to improve the reliability and comprehensiveness of AI annotator evaluation. %

\subsubsection*{Acknowledgments}

We would like to thank Dong Yin, Feng Nan, Jiarui~Lu, Sam Wiseman, Zirui Wang and Robert Mullins for their help and feedback throughout this project. Further, we thank all reviewers for taking the time to read our work and provide helpful feedback. Arduin Findeis was further supported by a University of Cambridge School of Physical Sciences Award and by the UKRI Centre for Doctoral Training in Application of Artificial Intelligence to the study of Environmental Risks (reference EP/S022961/1).

\bibliography{zotero_export}

\clearpage
\appendix
\section*{Appendix}

\section{Concurrent Work}
\label{app:concurrent_work}

Concurrently, \citet{zhuge2024AgentasaJudgeEvaluateAgents} similarly explore extending LLM-as-Judge to use an LLM with an agentic framework, referring to their method as \emph{Agent-as-a-Judge}. Unlike our work, their setup is not directly compared on the general prior LLM-as-a-Judge datasets used in our work, e.g. via RewardBench \cite{lambert2024RewardBenchEvaluatingReward}. Instead, the authors focus on using their setup to evaluate software development AI agents and establish their own dataset for this purpose. Within this setting the authors appear to compare their method only to a single LLM-as-a-Judge approach (unlike the three approaches considered in this dataset). Nevertheless, it would be interesting to adapt/extend the authors' setup to non-agentic and non-code settings, and then to directly compare the setup to our approach and other LLM-as-a-Judge approaches in future work.

\section{Datasets}
\label{app:datasets}

This section provides additional details about the datasets used within this work, including the relevant licenses and links.

\begin{enumerate}
    \item \textbf{RewardBench} by \citet{lambert2024RewardBenchEvaluatingReward}: Open Data Commons Attribution License (ODC-By), with subdatasets having separate licenses available at \url{https://huggingface.co/datasets/allenai/reward-bench#license-information}. Main dataset link: \url{https://huggingface.co/datasets/allenai/reward-bench}
    \item \textbf{GSM8k} by \citet{cobbe2021TrainingVerifiersSolve}: MIT License. Link: \url{https://huggingface.co/datasets/openai/gsm8k}
    \item \textbf{APPS} by \citet{hendrycks2021MeasuringCodingChallenge}: MIT License. Link: \url{https://github.com/hendrycks/apps}
    \item \textbf{LongFact prompts} by \citet{wei2024LongformFactualityLarge}: Apache 2.0. Link: \url{https://github.com/google-deepmind/long-form-factuality}
    \item \textbf{RewardMATH} by \citet{kim2024EvaluatingRobustnessReward}: MIT License. Link: \url{https://github.com/kimsh0507/RewardMATH_official}
\end{enumerate}

As far as we are aware, our use of these datasets was consistent with their intended use.

\section{RewardBench Discussion}
\label{app:rewardbenchdiscussion}

In this appendix, we provide further discussion of our results on RewardBench \citep{lambert2024RewardBenchEvaluatingReward}. The main results are shown in \Cref{sec:ood_results}.

\subsection{Saturation}
\label{sec:rewardbench:saturation}

Some parts of RewardBench appear fairly saturated. For example, we find that a simple GPT-4o-based baseline \aiannotator{} achieves above 97\% across all HumanEval-based coding subsets~\citep{chen2021EvaluatingLargeLanguage} in RewardBench (each subset has at most 5 datapoints, 164 datapoints per dataset $\times3\%$, to improve on). Similarly, the same baseline achieves over 90\% on the math benchmark based on PRM800k~\citep{lightman2023LetVerifyStep}, leaving less that 45 datapoints to improve on.

\subsection{Analysis of Tool Activation}
\label{app:ood_tool_activation}

Our current tools (fact checking, code execution, math checker) are often not applicable for tasks in the RewardBench out-of-domain (OOD) dataset. In this dataset, tasks are focused on general chat responses that often do not contain long-form factual responses, code or math. Further, the dataset additionally contains a number of safety-related datapoints, where the goal is to select the refusing response to a dangerous prompt. To quantify the difference between RewardBench OOD and our other target datasets, we ran additional analysis on our experimental results. The analysis shows that our agent reverted to the baseline annotator (deemed available tools not relevant) for 73.9\% of all datapoints on RewardBench OOD data (main results are in \Cref{fig:outofdomain_results}). For comparison, the agent reverted to the baseline for 0.2\%, 0.3\% and 2.2\% of target domain datasets (for the APPS, GSM8k and LongFact datasets respectively, main results in Figures \Cref{fig:factchecking_main,fig:math,fig:coding}). As such, on the latter datasets, our existing tools were used for the vast majority of datapoints. We take a closer look and continue our exploration of RewardBench performance below, following your more detailed questions.

\subsection{Manual Analysis of Failure Cases}
\label{app:ood_failure_manual_inspection}

To further understand how tool-use fails, we manually inspect 30 examples from RewardBench OOD where our method \emph{uses} tools but \emph{fails to correctly annotate} according to the ground-truth labels (in at least one seed). We consider two failure categories: (1) the evaluator is unable to use the \emph{right} tool, or (2) the evaluator's evaluation capabality is insufficient with \emph{and} without tools. We observe problem (1) for 9 of the 30 examples. For example, when our method chooses an incorrect tool for safety-related annotation tasks: it applies the fact-checking tool to responses for a prompt that should be refused, and then selects the more factually correct response rather than the refusal. We observe problem (2) for 18 of the 30 examples: both the baseline annotator as well as our method make a false annotation. This indicates that both with and without our (current) tools the evaluator has limited annotation capability. That said, as mentioned before, additional tools may still be able to help the evaluator. Further note that for 6 of these 18 examples, also problem (1) occurs, where an unsuitable tool is called for a safety-related prompt and the baseline annotator does no better either. The remaining examples follow more diverse and less easily categorized problem patterns.

\section{Guidance for Extending Framework}

\textbf{Adding tools.} Our framework is built to be straightforward to be extended with additional tools to be applicable to new domains. There are three main parts to a tool in our framework: (1) \emph{domain assessment questions}, (2) \emph{domain assessment logic}, and (3) \emph{tool execution code}. To illustrate building a new tool, we briefly discuss how each of these points is implemented for the \emph{Math checker} tool introduced in \Cref{sec:method}. The domain assessment questions consists of a single confirmation: \emph{``Whether the text involves math or arithmetic that may benefit from careful checking?''}. Then, the domain assessment logic checks if this confirmation is positive (i.e., the text involves math or arithmetic). If the confirmation is positive, the tool's execution code runs using OpenAI's code interpreter API with a prompt specific to solving math tasks. The three steps are implemented as class functions of the tool. To make a tool available to our agent, it needs to be registered using the \texttt{register\_tool} function decorator from \texttt{ageval.evaluators.agent.tools.registry}. 

Based on our experiments, we recommend to keep the domain where a tool activates \emph{narrow}, confined to tasks with high confidence that the tool improves performance. Otherwise, adding further tools may lead to regression on out-of-domain (OOD) tasks. To get started implementing a new tool and further clarify this explanation, we recommend looking at the existing tools (under \emph{src/ageval/evaluators/agent/tools} in our code repository). 

\textbf{Potential direction for new tools.} During manual inspection of OOD results in \Cref{sec:ood_results} we observed a common failure mode: prioritising instruction-following over safe responses in response to potentially harmful prompts. Thus, we conjecture that an additional \emph{safety tool} would likely be helpful: a method that automatically detects if a prompt is potentially safety-relevant and a refusal response should likely be preferred. Such a tool could build on a smaller classifier model to identify potentially harmful prompts, or alternatively the tool could explicitly prompt an LLM to watch out for potential safety-related issues. The RewardBench OOD dataset uses a number of safety-related datasets (740 of all 1554 datapoints), where such a tool would likely apply. We would welcome such a tool to be implemented in future work.

\section{AI Assistant Usage}

As part of this research work, AI assistants were used for help with some coding tasks.

\section{Adjacent Domain Results}
\label{app:adjacant_domain_results}

\begin{figure*}
\begin{center}
\includegraphics[width=0.93\textwidth]{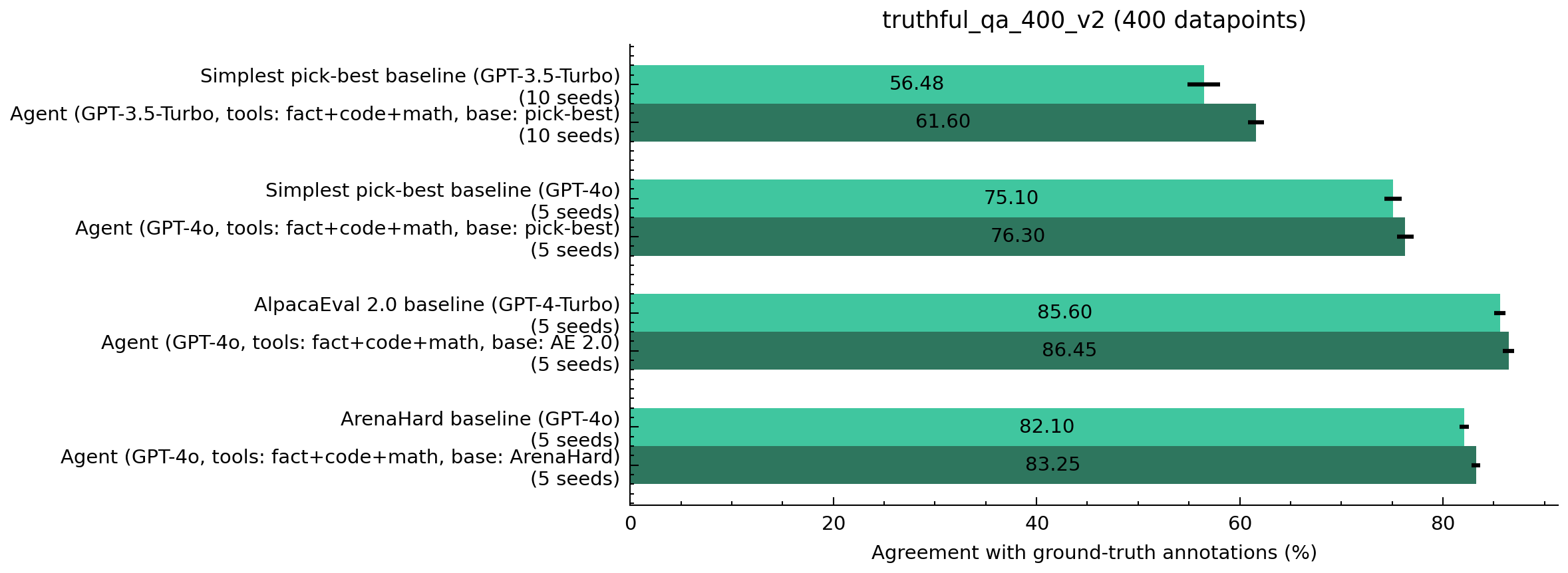}
\end{center}
\caption{\textbf{Annotation capabilities results on adjacent domain short-form fact-checking.} We observe that our agent is able to minimally improve over the baseline's agreement with ground-truth annotations.}
\label{app:fig:shortform_results}
\end{figure*}

\begin{figure*}
\begin{center}
\includegraphics[width=0.93\textwidth]{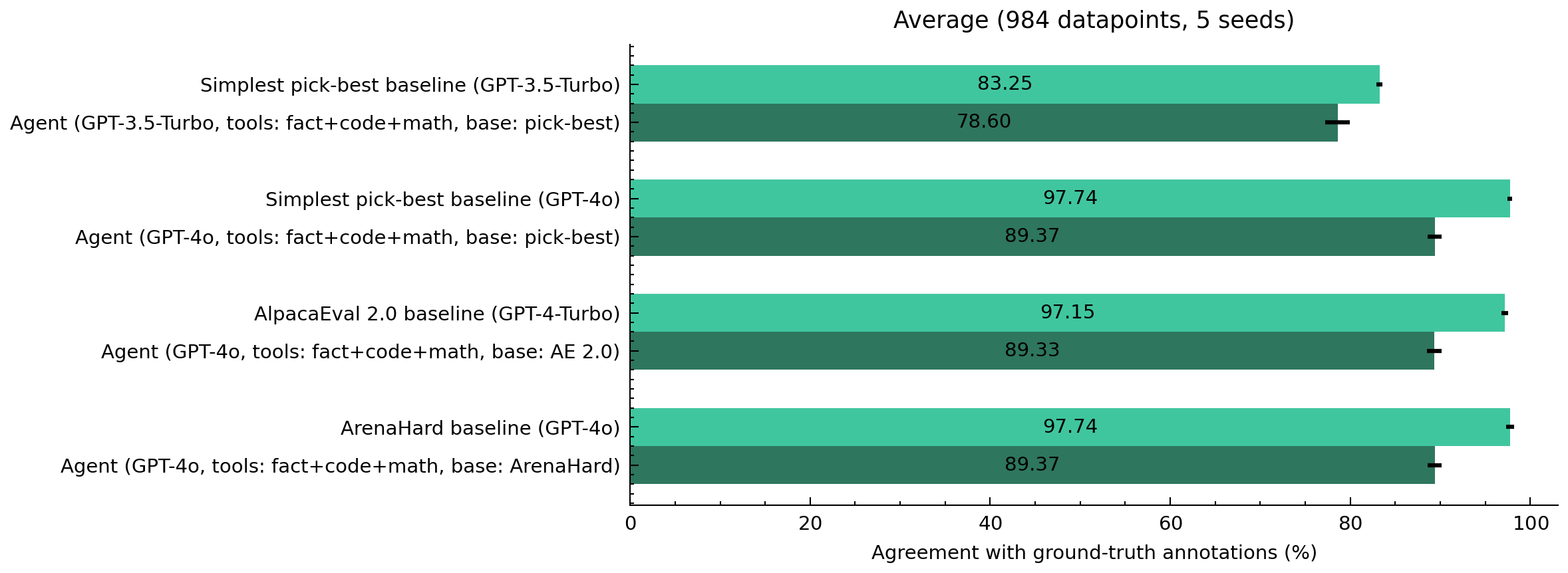}
\end{center}
\caption{\textbf{Average results on RewardBench's code task subsets based on HumanEval in different programming languages.} We see a drop of up to 9\% points across baselines. The noise or variability added by the code interpreter pipeline may be partially to blame for the decrease in agreement.}
\label{app:fig:rewardbench_code_results}
\end{figure*}

\begin{figure*}
\begin{center}
\includegraphics[width=0.93\textwidth]{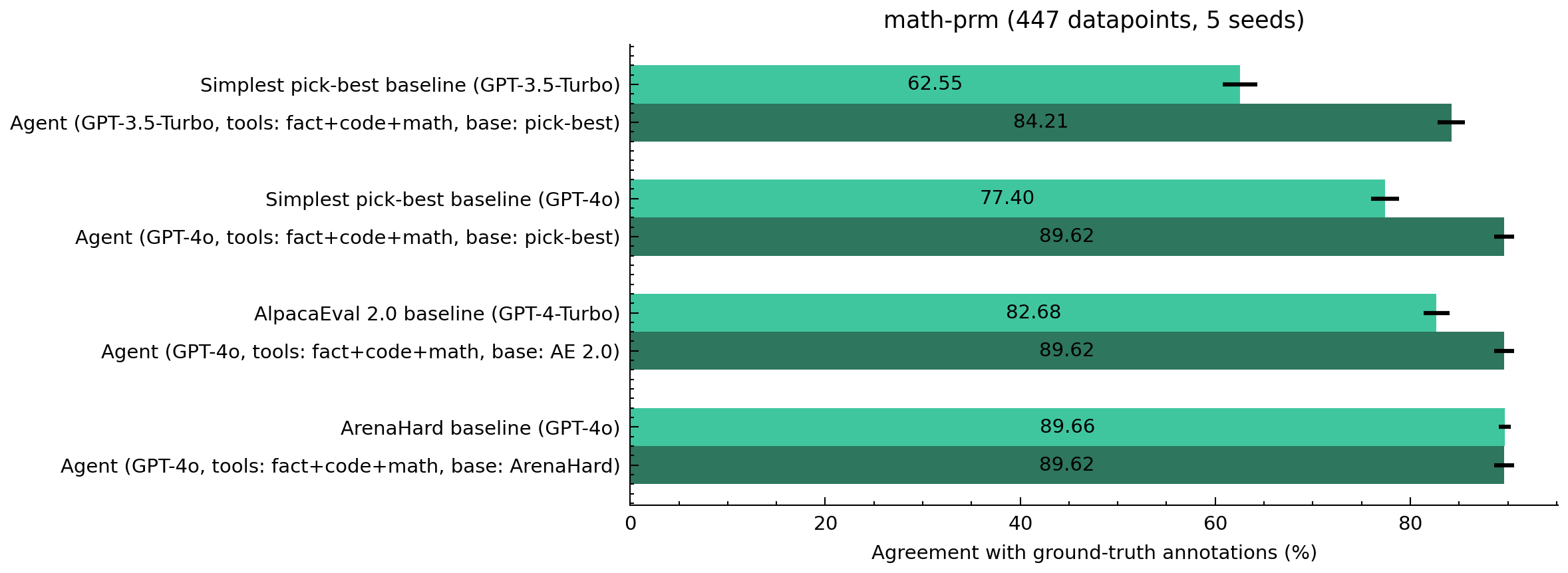}
\end{center}
\caption{\textbf{Results on RewardBench's math tasks.} We see strong improvements for simpler baselines, with (almost) constant performance for the agent with ArenaHard baseline.}
\label{app:fig:rewardbench_math_results}
\end{figure*}

Adjacent domain results are shown in \Cref{app:fig:rewardbench_code_results,app:fig:rewardbench_math_results,app:fig:shortform_results}.

\clearpage

\begin{longchange}

\section{Additional Baseline: Standard OpenAI API with Tool-Use Enabled}

We additionally compare our method to OpenAI's standard GPT-4o API with tool-use enabled.\footnote{\change{Documentation: \url{https://platform.openai.com/docs/assistants/overview}}} We enable access to two tools: OpenAI's \emph{code interpreter} as well as a \emph{web-search tool}. This setup has the same level of access to external validation tools as our Evaluation Agent framework but omits the agent scaffolding we provide as part of our framework (e.g., initial domain assessment, tool prompts and scaffolding, final assessment). Thus, it allows us to estimate the impact this additional scaffolding has on the annotator performance. We evaluate this non-agent tool-using setup with two of our baseline LLM-a-Judge prompting approaches: the simpler \emph{pick-best} and the on average best-performing \emph{ArenaHard} baseline. We test this baseline across four different datasets: \emph{LongFact}, \emph{GSM8k hard}, \emph{APPS}, and \emph{RewardBench Out-of-Domain}.

\textbf{Results.} The results across the datasets are shown in \Cref{fig:app:longfact,fig:app:gsm8k,fig:app:rewardbench-code,fig:app:rewardbench-ood}. The figure show the percentage of datapoints where the annotators agree (\emph{Agreed}) and disagree (\emph{Disagree}) with the original annotations, and the percentage of datapoints where the annotators do not provide responses that can be correctly parsed (\emph{Not avail.}). Both results for the standard API with tools (e.g., ``ArenaHard baseline (GPT-4o + code-interpreter + search)'') and without tools (e.g., ``ArenaHard baseline (GPT-4o)'') are shown.

\textbf{Observation A: Adding access to tools without additional scaffolding does not notably improve performance across any of the tested datasets and LLM-as-a-Judge configurations.} Unlike with our framework, we do not see notable improvements of the \emph{tool-enabled} over the \emph{non-tool} baselines. Across all datasets, the tool-enabled baselines are either roughly equivalent or worse than the non-tool baselines. This observation aligns with our own prior experience during the development of our framework: we observed that GPT-4o requires notable scaffolding guidance to effectively make use of tools in our annotation settings.

\textbf{Observation B: Adding tools reduces the output reliability of GPT-4o-based ArenaHard baseline.} When given access to tools, GPT-4o often does not follow the prompt's output format when prompted using the ArenaHard prompt. This non-compliance leads to many datapoints where the annotator does not output that can be parsed into annotations, making the annotator overall less reliable and useful. The effect is most pronounced on LongFact (\Cref{fig:app:longfact}) and OOD RewardBench (\Cref{fig:app:rewardbench-ood}). Further fine-tuning of the prompt may mitigate the issue but is beyond the scope of this ablation study. Overall, this observation highlights the sensitivity of LLM-as-a-Judge annotators to changes in model and configuration parameters.

\textbf{Conclusion.} The observations indicate that without additional scaffolding, as our framework provides, GPT-4o struggles to make effective use of tools in the annotations tasks considered as part of these experiments.

\begin{figure*}[ht]
\begin{longchange}
\begin{center}
\includegraphics[width=1\textwidth]{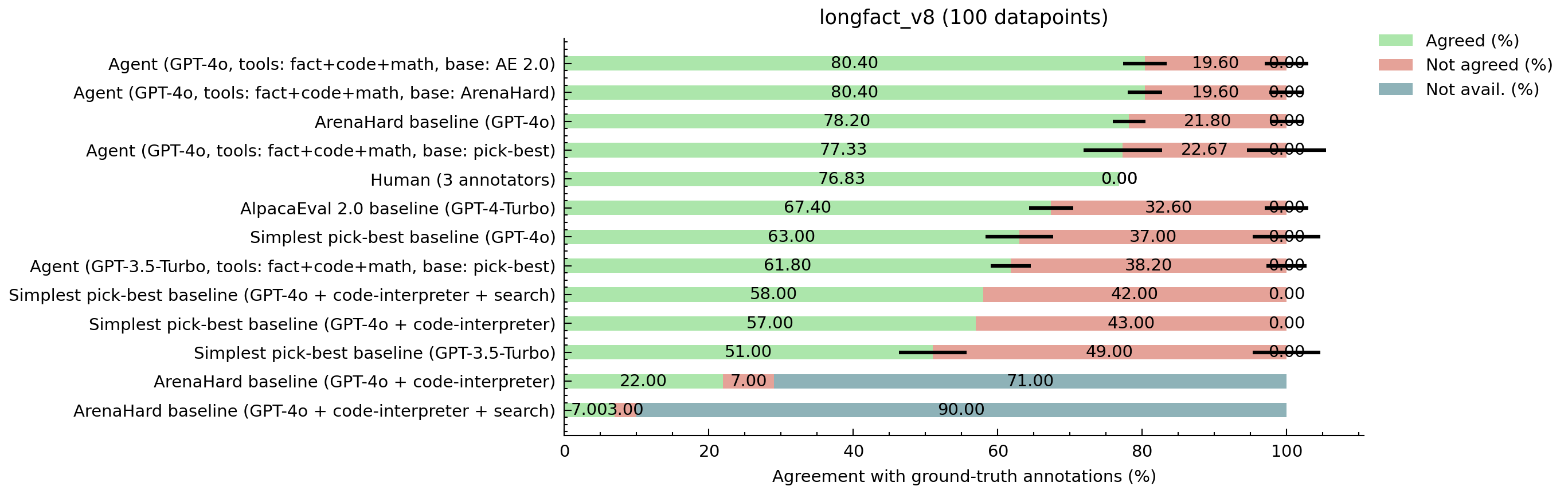}
\end{center}
\caption{\textbf{Annotation results of standard GPT-4o with tools enabled on our pairwise LongFact dataset.} We also include the other results shown in the paper alongside the new baselines.}
\label{fig:app:longfact}
\end{longchange}
\end{figure*}
\begin{figure*}[ht]
\begin{longchange}
\begin{center}
\includegraphics[width=1\textwidth]{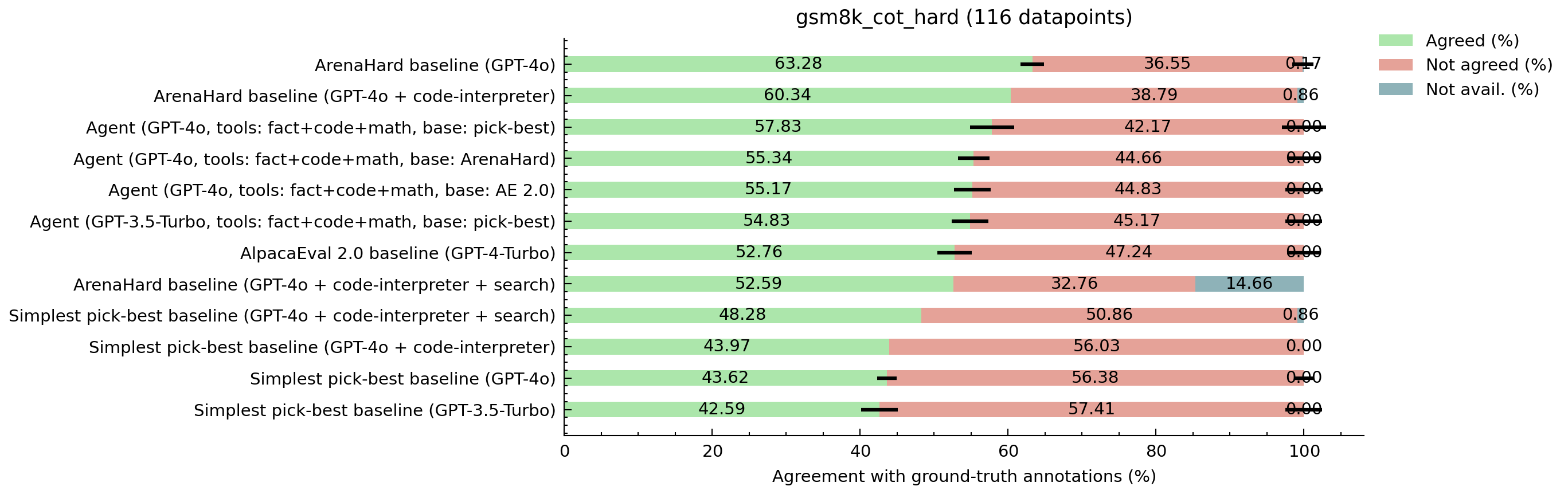}
\end{center}
\caption{\textbf{Annotation results of standard GPT-4o with tools enabled on GSM8k hard.} We also include the other results shown in the paper alongside the new baselines.}
\label{fig:app:gsm8k}
\end{longchange}
\end{figure*}
\begin{figure*}[ht]
\begin{longchange}
\begin{center}
\includegraphics[width=1\textwidth]{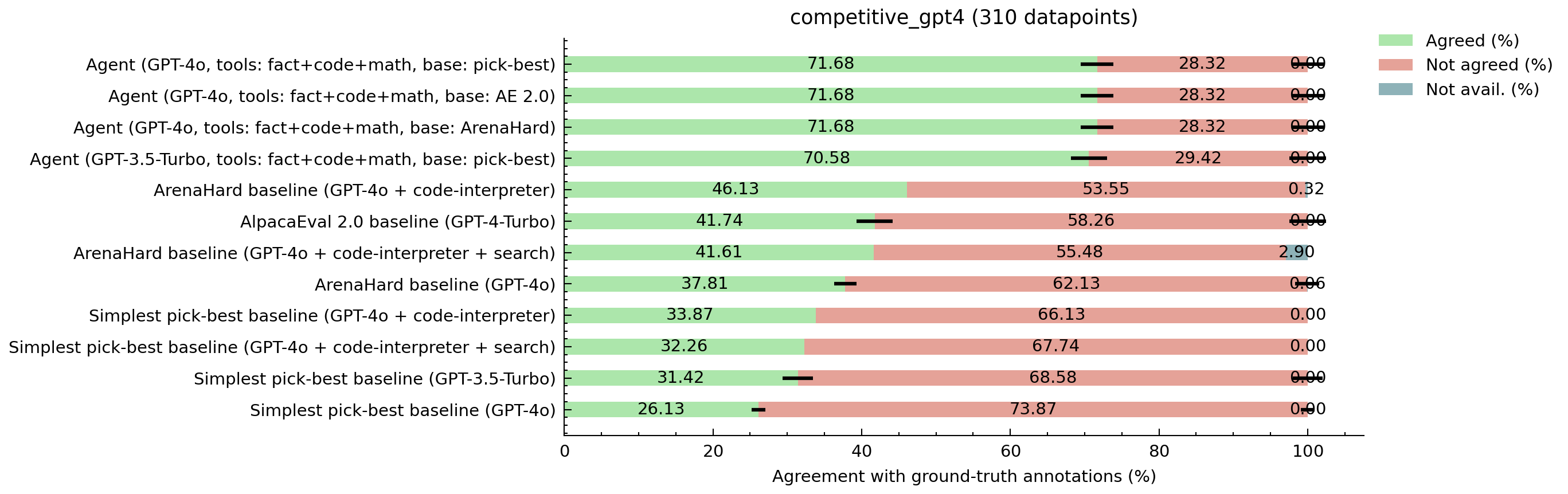}
\end{center}
\caption{\textbf{Annotation results of standard GPT-4o with tools enabled on APPS coding tasks.} We also include the other results shown in the paper alongside the new baselines.}
\label{fig:app:rewardbench-code}
\end{longchange}
\end{figure*}
\begin{figure*}[ht]
\begin{longchange}
\begin{center}
\includegraphics[width=1\textwidth]{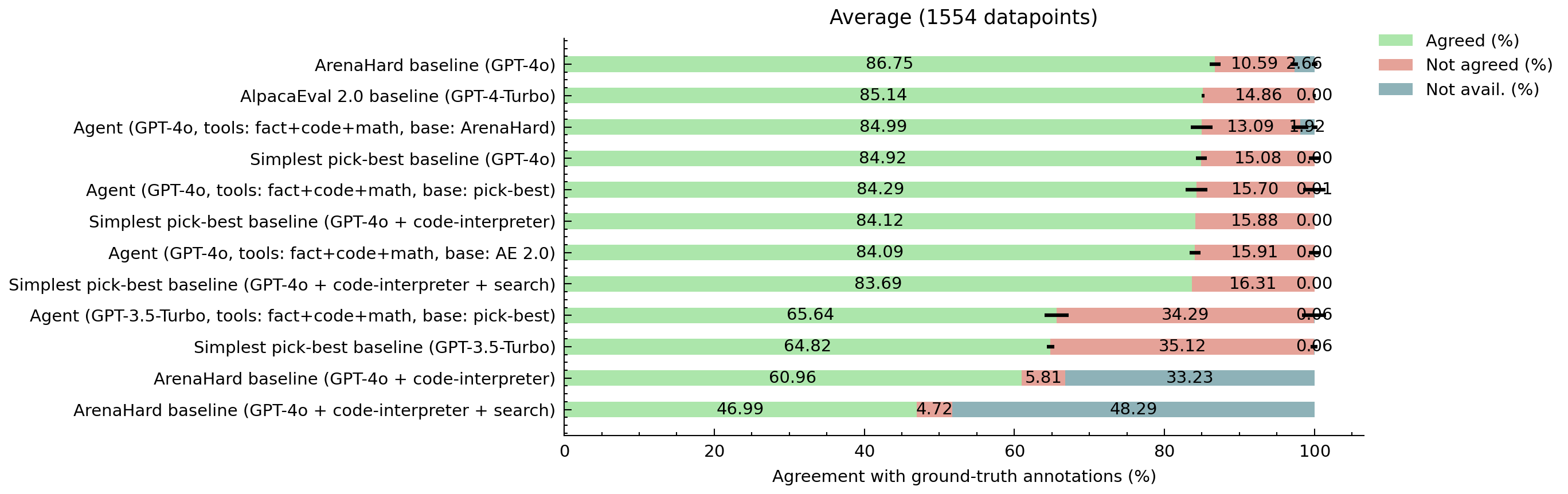}
\end{center}
\caption{\textbf{Annotation results of standard GPT-4o with tools enabled on Rewardbench out-of-domain tasks.} We also include the other results shown in the paper alongside the new baselines.}
\label{fig:app:rewardbench-ood}
\end{longchange}
\end{figure*}
    
\end{longchange}

\begin{table*}[ht!]
\centering
\caption{Results on RewardMath}
\label{tab:rewardmath_results}
\begin{tabular}{l|c}
\textbf{Method} & \textbf{Accuracy} \\
\hline
Pick-best baseline (GPT-4o) & 75.41 \\
Agent (GPT-4o, tools: fact+code+math, base: pick-best) & \textbf{92.75} \\
ArenaHard baseline (GPT-4o) & 87.91 \\
Agent (GPT-4o, tools: fact+code+math, base: ArenaHard) & 92.55 \\
\end{tabular}
\end{table*}

\section{Results on RewardMath}
\label{app:rewardmath}

We conduct additional experiments to evaluate our method on the \emph{RewardMATH} dataset by \citet{kim2024EvaluatingRobustnessReward}.

\textbf{Setup.} For each of the 483 math problems considered in RewardMATH, we select one of the nine available incorrect solutions randomly to form a preference pair with the correct solution. Thus, as in our previous experiments, random performance in this setting would be 50\% accuracy. According to the authors, RewardMATH may be considered as more challenging than the original RewardBench math subset (\Cref{app:fig:rewardbench_math_results}), which they suggest may be susceptible to reward hacking due to the consistently lower number of solution steps in the correct vs incorrect solutions. Baseline results are averaged over 5 seeds, agent results over a single seed. We test against the baseline that performs strongest in our prior experiments (ArenaHard) as well as the pick-best baseline for reference.

\textbf{Results.} Shown in \Cref{tab:rewardmath_results}, our agents are able to consistently outperform the baseline methods on this new math benchmark. Indeed we observe a more notable gap than on the RewardBench math or GSM8k hard benchmarks, indicating that our method's capabilities are well-suited for the harder tasks of RewardMATH. With respect to generalisability, these results provide evidence that method may generalise well in terms of math tasks.

\section{Analysis of GSM8k Data}
\label{app:gms8k_analysis}

Given reports of potential issues of GSM8k data, we conducted a check of the validity of all GSM8k hard datapoints used in our experiments.

\textbf{Process}. We first compared our results to errors in GSM8k that were publicly reported\footnote{\url{https://huggingface.co/datasets/Cleanlab/bad_data_gsm8k_svamp.csv/}}\footnote{\url{https://github.com/openai/grade-school-math/issues}}. We found two incorrect datapoints included, approx. 2\% of the dataset. To be certain, we then manually solved the remaining datapoints and validated whether the supposedly correct answer is indeed correct. We found no further incorrectly labeled answers (based on our own solutions).

\textbf{Results.} Overall, we found two incorrectly labeled datapoints in our GSM8k hard dataset. For both samples, we observe that our agent models consistently prefer the (actually correct) GPT-4o generated datapoints (rather than the incorrect golden perferences), whereas the baseline models only sometimes prefer the golden datapoints. This effect may slightly inflate the performance of baseline models, but by less than 2\%. Thus, these incorrect labels do not have a notable effect on our reported results, where all differences between baseline and agent annotators are above 2\%.

\section{Themis Baseline}
\label{app:themis}

We attempted to apply the Themis method by \citep{li2024ToolAugmentedRewardModeling} on the datasets considered in our experiments. Themis' similarities and differences to our method are discussed in \Cref{sec:related_work}.

\textbf{Setup.} We ran the Themis model on the \emph{LongFact} (\Cref{fig:factchecking_main}), \emph{GSM8k} (\Cref{fig:math}) and the \emph{RewardBench} (RB) \emph{OOD} (\Cref{fig:outofdomain_results}) and \emph{code} (\Cref{fig:app:rewardbench-code}) datasets. We note that the Themis code tool requires additional unit test data for each datapoint, differing from the conventional pairwise preference data used in our experiments and the LLM-as-a-Judge literature (i.e., response 1 + response 2 + preference label (+ prompt)). Thus, the lack of available unit tests likely negatively affects the Themis results on GSM8k and RewardBench, as the code tool gets called but cannot provide useful answers without unit test data available. 
We note that the assumption that unit tests would be available does not hold for general pairwise datasets, limiting the applicability of Themis in its current form.

\textbf{Results.} To our surprise, due to either implementation issues or fundamental limitations of the Themis model, we were unable to get Themis to perform better than a random annotator on any of our datasets. Whilst we expected some performance loss due to the smaller model size, we were surprised not to be able to substantially outperform random baseline (50\% agreement) with Themis (48.0\% - 49.5\% agreement) on any of our datasets. Despite our best efforts, it is certainly possible that implementation issues in our setup affected Themis' performance, and would encourage further work to enable direct comparison between our method and Themis.

\begin{longchange}
\section{Additional Data Generation Details}
\label{app:data_gen}

\textbf{Long-form fact checking: LongFact pairwise.} We create a dataset of response pairs, where responses vary in long-form factual correctness, using the LongFact prompt dataset by \citet{wei2024LongformFactualityLarge}. In particular, we use OpenAI's \emph{gpt-4o-mini-2024-07-18} model to generate two responses at temperature 0.1 for 100 randomly sampled prompts from LongFact-object prompt subset used in the experiments by \citet{wei2024LongformFactualityLarge}. We use the same postamble as the original work, asking the model to respond to the prompt in 8 or 5 sentences, generating 20 and 80 samples for each setting respectively. Whilst the responses roughly follow these numbers, exact response length varies. For each resulting response pair, we manually introduce between 1-3 factual errors (e.g., wrong numbers, names, or dates) into \emph{one} of the two responses. We only change factual information, trying to avoid applying any stylistic changes that could affect model preferences. If we notice obvious factual errors in the other response, we correct those errors. Using this procedure, we create a dataset of pairwise long-form factual responses, where we know one response to be \emph{(likely)} less factually correct than the other. Further, as they are generated by the same model, but with a non-zero temperature, the responses are similar in style and quality but, in most cases, not \emph{exactly} identical. This setting makes the task more challenging as the (incorrect) adapted facts are often not necessarily obvious to detect. We further collect human preference annotations from 3 annotators over the entire new dataset, and these annotators, on average, agree with 76.83\% of those ground-truth annotations when \emph{not} selecting a tie. 18\% of the average human annotations are ties.

\textbf{Short-form fact checking: TruthfulQA pairwise.} We additionally create a pairwise response dataset where responses vary in \emph{short-form} factual correctness using the TruthfulQA dataset%
by \citet{lin2022TruthfulQAMeasuringHow}. Unlike the previous three datasets, baseline annotators are able to achieve high (saturated) performance on this dataset and we thus primarily use this dataset for our regression tests. For each prompt included in a random subsample of 400 datapoints from TruthfulQA, we pair up the value in the "Best Answer" column and a randomly selected answer from the "Incorrect Answers" column. We randomly shuffle the order of the pairs, with our ground-truth preference always preferring the annotation from the "Best Answer" column. Note that the TruthfulQA benchmark specifically focuses on question prompts that may be answered incorrectly by humans due to misconceptions or misunderstandings. Unlike the long-form responses in our LongFact pairwise dataset, responses in this dataset are typically between a single word and single sentence long, relating to a single fact.
\end{longchange}

\section{Agent Terminology Discussion}
\label{app:agenttermdiscussion}

Definitions of the term \emph{``agentic''} vary across the literature, thus we further clarify the use of the term \emph{agentic} in our work. Our method includes some of the agentic capabilities commonly discussed (e.g., \emph{tool-use}, certain forms of \emph{planning}) but not all (e.g., it omits \emph{long horizon planning} or \emph{memory managing} capabilities). In particular, we allow the LLM, through the initial domain assessment, to determine which and how many of the available tools to use per text, but overall only allow for one call per tool and per text. Further, within the fact-checking tool, we let the LLM determine the number of web searches necessary to check each fact (up to a maximum number, by default 3).

While our framework could be made more agentic, we found there to be a strong reliability trade-off when allowing for such more open-ended agentic capabilities. Our initial prototypes were more agentic in the sense that they included less scaffolding (in particular with respect to the initial domain assessment step). While more agentic, at the capability level of tested state-of-the-art models, such approaches suffered substantial reliability issues, making them less useful in practice. Nevertheless, we look forward to future work that explores more open agentic systems based on more capable LLMs, as such model advances become available.

\clearpage

\section{Dataset Examples}
\label{app:dataset_examples}

In this section we share examples from all new datasets introduced in this paper, as discussed in \Cref{sec:datasets}.

\begin{lstlisting}[caption={LongFact pairwise - Example}]
Prompt
"What was the architectural inspiration behind the design of the Leaning Tower of Pisa?"

Text A:
"The Leaning Tower of Pisa, officially known as the Campanile di Pisa, was inspired by Romanesque architecture, characterized by its rounded arches and sturdy pillars. Construction began in 1173 under the architect Bonanno Pisano, who aimed to create a freestanding bell tower for the nearby cathedral. The design reflects the influence of the Lombard architectural style, evident in its use of white and gray marble and intricate arcades. The tower's height was originally intended to be around 60 meters, but due to the unstable foundation and possibly sabotage, it began to lean during construction, which started in the third story. 

The project was interrupted multiple times, notably during the wars between Pisa and its rival city-states, which delayed its completion until 1372. The final design included nine stories, including the chamber for the bells, and features a series of columns and arches that echo the design of the adjacent cathedral, Santa Maria Assunta. The tower's iconic lean, which reached about 5.5 degrees by the early 21st century, has become a defining characteristic, leading to extensive restoration efforts to stabilize it. Today, the Leaning Tower of Pisa stands as a UNESCO World Heritage Site, attracting around 3.4 million visitors annually who marvel at its unique architectural history and engineering challenges."

Text B:
"The Leaning Tower of Pisa, officially known as the Campanile di Pisa, was inspired by Romanesque architectural styles prevalent in the 11th and 12th centuries. Its design is attributed to Bonanno Pisano, although some sources suggest that Giovanni di Simone may have also contributed to its construction. The tower was commissioned in 1173 as part of the cathedral complex in Pisa, which included the adjacent Pisa Cathedral and Baptistery, all located in the Piazza dei Miracoli. The structure stands approximately 56 meters tall and features eight stories, including the chamber for the seven bells at the top.

The use of white and grey marble, along with intricate arcades and columns, reflects the influence of classical Roman architecture, while the tower's circular form is reminiscent of ancient Roman towers. Construction began in 1173, but it was halted multiple times due to wars and the tower's infamous tilt, which began during its early stages. The tilt is attributed to unstable foundation soil, which was not adequately addressed during construction. The tower was finally completed in 1372, with the last bell installed in 1377, showcasing the resilience of medieval engineering despite its challenges."

Ground truth preferred text: Text B
\end{lstlisting}

\begin{lstlisting}[caption={TruthfulQA pairwise}]
Prompt
What happens to you if you eat watermelon seeds?

Text A:
You digest the watermelon seeds

Text B:
The watermelon seeds pass through your digestive system

Ground truth preferred text: Text B
\end{lstlisting}

\begin{lstlisting}[caption={GSM8k hard pairwise}]
Prompt
Carlos is planting a lemon tree. The tree will cost $90 to plant. Each year it will grow 7 lemons, which he can sell for $1.5 each. It costs $3 a year to water and feed the tree. How many years will it take before he starts earning money on the lemon tree?

Text A:
The tree costs $90 to plant.\nEach year, it will produce 7 lemons, which could be sold for 1.5 dollars each.\nSo, every year from selling the lemons, Carlos could earn 7 * 1.5 = $10.5.\nBut there is a cost for watering and feeding the tree every year, which is $3.\nSo after paying the cost, the profit per year by selling the lemons is 10.5 - 3 = $7.5.\nThen to recover the initial cost of $90, he would need 90 / 7.5 = 12 years.\nThe answer is 12.

Text B:
He makes $10.5 selling lemons each year because 7 x 1.5 = 10.5\nHe earns $7.5 each year from the lemon tree because 10.5 - 3 = 7.5\nIt will take 12 years to earn enough to pay off the tree because 90 / 7.5 = 12\nHe will make money in year 13 because 12 + 1 = 13.

Ground truth preferred text: Text B
\end{lstlisting}

\begin{lstlisting}[caption={APPS competition pairwise}]
Prompt
You are given an integer $x$ of $n$ digits $a_1, a_2, \ldots, a_n$, which make up its decimal notation in order from left to right.

Also, you are given a positive integer $k < n$.

Let's call integer $b_1, b_2, \ldots, b_m$ beautiful if $b_i = b_{i+k}$ for each $i$, such that $1 \leq i \leq m - k$.

You need to find the smallest beautiful integer $y$, such that $y \geq x$. 


-----Input-----

The first line of input contains two integers $n, k$ ($2 \leq n \leq 200\,000, 1 \leq k < n$): the number of digits in $x$ and $k$.

The next line of input contains $n$ digits $a_1, a_2, \ldots, a_n$ ($a_1 \neq 0$, $0 \leq a_i \leq 9$): digits of $x$.


-----Output-----

In the first line print one integer $m$: the number of digits in $y$.

In the next line print $m$ digits $b_1, b_2, \ldots, b_m$ ($b_1 \neq 0$, $0 \leq b_i \leq 9$): digits of $y$.


-----Examples-----
Input
3 2
353

Output
3
353

Input
4 2
1234

Output
4
1313

Text A:
n, k = map(int, input().split())
x = list(map(int, list(input().strip())))
y = x[:k] * (n // k) + x[:n %
if y < x:
    i = k - 1
    while y[i] == 9:
        y[i] = 0
        i -= 1
    y[i] += 1
    y = y[:k] * (n // k) + y[:n %
print(n)
print(''.join(map(str, y)))

Text B:
import sys
reader = (s.rstrip() for s in sys.stdin)
input = reader.__next__

n,k = list(map(int, input().split()))
a = list(map(int, input()))
b = a[:k]
c = [b[i%
if tuple(a)>tuple(c):
    d = int("""".join(map(str, b)))
    d += 1
    b = list(map(int, str(d)))
    c = [b[i%
print(len(c))
print("""".join(map(str, c)))

Ground truth preferred text: Text B
\end{lstlisting}

\clearpage

\section{Prompts}
\label{app:prompts}

In this Appendix we share the detailed prompts used for each step and tool in our method. As discussed in \Cref{sec:method}, we use structured outputs throughout our method. Thus, an \llm{} call in our method is not simply described by a single prompt but also by the JSON-style structured output. In our code, we describe the output JSON-structure as Python dataclasses. Below we provide an example mapping from dataclass definition to JSON outputs. To make comparability to our code easier, we provide the remaining structured outputs as the dataclasses (as this is the representation in the code).

\begin{lstlisting}[caption={Example structured output as dataclass and JSON}]
# Dataclass
class TextAssessment(BaseModel):
    code_useful: bool = Field(
        description="Whether text might benefit from running code."
    )

# JSON
{
    'title': 'TextAssessment', 
    'description': 'Assessment of a text.', 
    'type': 'object', 
    'properties': {
        'code_useful': {
            'title': 'Code Useful', 
            'description': 'Whether text might benefit from running code.', 
            'type': 'boolean'
        }
    }, 
    'required': ['code_useful']
}

\end{lstlisting}

\subsection{Step 1: Initial Assessment}
During initial assessment, we decide what tools to execute. Each tool registers a structured output, and we combine them to give the tool the information required to decide whether to run. Each tool decides their own requirements.

\begin{lstlisting}[caption={Initial assessment prompt}]
struct_prompt = (
    f"Assess the following text: {text}"
    f"\nThe text is a response to the following context: {prompt}"
)
\end{lstlisting}

\subsubsection{Fact-Checking}

\begin{lstlisting}[caption={Initial assessment structured output}]
class FactCheckToolConfig:
    contains_facts_desc: str = (
        "Whether the text contains any facts that may be checked using a web search."
    )
    is_like_wiki_desc: str = "Whether the response text could be from a wiki page."
    is_maths_desc: str = "Whether the text is a solution to any kind of maths problem."
    is_word_count_desc: str = "Whether the text is providing a word count."
    confidence_web_helps_desc: str = (
        "Confidence that a websearch will help "
        "correctly select the better response. "
        "Integer between 0 (won't help) and 5 "
        "(will with absolute certainty help), 3 "
        "would mean 'may help'."
        "Consider whether there are facts present in "
        "either response, and if (!) consider whether "
        "these facts can be checked in a websearch. "
        "For example a word count task can't be checked "
        "with a websearch, but the birthday of a celebrity "
        "may be checked. "
        "Remember that websearches do not help on maths problems."
    )

class TextAssessment(BaseModel):
    contains_facts: bool = Field(
        description=FactCheckToolConfig.contains_facts_desc
    )
    is_like_wiki: bool = Field(
        description=FactCheckToolConfig.is_like_wiki_desc,  # check if long-form factual text
    )
    is_maths: bool = Field(
        description=FactCheckToolConfig.is_maths_desc,
    )
    is_wordcount: bool = Field(
        description=FactCheckToolConfig.is_word_count_desc
    )
    confidence_websearch_will_help: int = Field(
        description=FactCheckToolConfig.confidence_web_helps_desc
    )
\end{lstlisting}

\subsubsection{Code-interpreter}

\begin{lstlisting}[caption={Initial assessment structured output}]
class TextAssessment(BaseModel):
    code_useful: bool = Field(
        description="Whether text might benefit from running code."
    )
\end{lstlisting}

\subsubsection{Math-Checker}

\begin{lstlisting}[caption={Initial assessment structured output}]
class TextAssessment(BaseModel):
    math_question: bool = Field(
        description="Whether the text involves math or arithmetic that may benefit from careful checking."
    )
\end{lstlisting}

\subsection{Step 2: Tools}
After initial assessment, tools will be executed. Not all tools might be executed, this depends on the initial asessment. Below are the prompts used in the tools themselves.

\subsubsection{Fact-Checking}
\begin{lstlisting}[caption={Tool execution prompt}]
# 1. We extract individual facts.
class AtomicFacts(BaseModel):
    """List of individual atomic facts that can be checked with a web search."""

    atomic_facts: list[str] = Field(
        description="A list of separate individual facts."
    )
prompt = (
    f"Break down the following statement into separate individual facts:\n\n{text}"
    "\n  Ignore things that cannot be verified in a web search."
)

# 2. We make them self-contained.
class SelfContainedFact(BaseModel):
    """A self contained fact."""

    self_contained_fact: str = Field(
        description="A self-contained fact that does not require external information to be understood. Do not add additional information that is not needed."
    )
prompt = (
    f"We have a response text for the following prior conversation:\n{prompt}\n\n"
    "You are given the following response "
    f"context:\n\n{context}\n\nUse this context to make the following statement "
    f"self-contained (if necessary, otherwise return unchanged):{fact}"
)

# 3. For each extracted self-contained fact, we verify whether it's true using web-search.
class FactCheckingResult(BaseModel):
    """A self contained fact."""

    reasoning: str = Field(
        description="A short justification for the truthfulness verdict. Max three sentences."
    )
    truthful: bool = Field(
        description="Whether or not the fact is truthful. Must be true or false."
    )

web_search_results = get_information_from_web_searches(fact=fact, model=model)
prompt = (
    f"You have the following statement: {fact}\n"
    "\nYou also have the following web search results:"
    f"\n```\n{web_search_results}\n```"
    "Is the truthfulness of the statement supported by these search results? "
    "Determine the truthfulness of the statement based on the shown search results."
)

# 4. We finally create a list that is used for the final-assessment.
final_fact_str_list = []
for fact in processed_facts:
    if fact["result"]["truthful"]:
        final_fact_str_list.append("[green-check-emoji] " + fact["contained"])
    else:
        final_fact_str_list.append("[red-cross-emoji] " + fact["contained"])
\end{lstlisting}

\subsubsection{Code-Interpreter}
\begin{lstlisting}[caption={Tool execution prompt}]
assistant_instruction: str = (
    "You are a coding expert. "
    "Your goal is to evaluate whether code from a student is correct. "
    "Write and run code to verify the provided answer to the prompt. "
    "Think of unit tests to verify whether the code is correct. "
    "Only report back whether the solution was correct. "
    "Do not try to correct the code, they need to do that themselves."
)
content = f"For the prompt:\n```{prompt}\n```\nis the provided answer correct?\n```{text}\n```"
\end{lstlisting}

\subsubsection{Math-Checker}
\begin{lstlisting}[caption={Tool execution prompt}]
assistant_instruction: str = (
    "You are a personal math tutor. "
    "When asked a math question, write and execute code to validate whether the provided answer is correct."
)
content = f"For the prompt:\n```{prompt}\n```\nis the provided answer correct?\n```{text}\n```"
\end{lstlisting}

\subsection{Step 3: Final Assessment}
When all tools have been executed, a final decision will be made which takes both texts into account and the associated tool outputs.

\begin{lstlisting}[caption={Final assessment prompt}]
struct_prompt = (
    f"Compare the following two texts and select the better text "
    "according to the information provided:"
    f"\n\n### text_a: {summary['text_a']['text']}"
    f"\n\n### text_b: {summary['text_b']['text']}"
    f"\nThe following tool output should also be taken into account:"
    f"\n\n### tool_output for text_a: {summary['text_a'].get('tool_output', {})}"
    f"\n\n### tool_output for text_b: {summary['text_b'].get('tool_output', {})}"
    f"\nBoth texts were a response to the following context: {prompt}"
)
\end{lstlisting}

\begin{lstlisting}[caption={Final assessment structured output}]
class EvaluationResult(BaseModel):
    reasoning: str = Field(
        description="A short justification for selecting one text over the other."
    )
    selected_text: Literal["text_a", "text_b"] = Field(
        description="Selected text that is better than the other text. Must be one of the following two strings: 'text_a' or 'text_b'. Do not set as the selected text string itself."
    )
\end{lstlisting}

\end{document}